%% file: main.tex
\documentclass[letterpaper]{article} 
\usepackage[submission]{aaai25}  
\usepackage{times}  
\usepackage{helvet}  
\usepackage{courier}  
\usepackage[hyphens]{url}  
\usepackage{graphicx} 
\usepackage{booktabs}
\usepackage[table,xcdraw]{xcolor}
\usepackage{pifont}%
\newcommand{\xmark}{\ding{55}}%

\usepackage{natbib}  
\usepackage{caption} 
\usepackage{algorithm}
\usepackage{algorithmic}
\usepackage{subcaption}

\usepackage{newfloat}
\usepackage{listings}
\usepackage{graphicx}  
\usepackage{amsmath}   
\usepackage{textcomp}  
\usepackage{amsfonts}  
\usepackage{bm}        
\DeclareCaptionStyle{ruled}{labelfont=normalfont,labelsep=colon,strut=off} 
\floatstyle{ruled}
\newfloat{listing}{tb}{lst}{}
\floatname{listing}{Listing}
\usepackage{bibentry}

\def\showauthors@on{T}

\title{
CLIP meets DINO for Tuning Zero-Shot Classifier using 
\\
Unlabeled Image Collections
}

\author {
    Mohamed Fazli Imam\textsuperscript{\rm 1},
    Rufael Fekadu Marew\textsuperscript{\rm 1},
    Jameel Hassan\textsuperscript{\rm 1,2},
    \\
    Mustansar Fiaz\textsuperscript{\rm 3},
    Alham Fikri Aji\textsuperscript{\rm 1},
    Hisham Cholakkal\textsuperscript{\rm 1}
}
\affiliations {
    \quad \textsuperscript{\rm 1}Mohamed Bin Zayed University of  AI
    \quad \textsuperscript{\rm 2}The Johns Hopkins University
    \quad \textsuperscript{\rm 3}IBM Research\\
    \footnotesize \quad \texttt{mohamed.imam@mbzuai.ac.ae}
    \quad \texttt{rufael.marew@mbzuai.ac.ae}
    \quad \texttt{jameel.hassan@mbzuai.ac.ae}
    \\
    \quad \texttt{mustansar.fiaz@ibm.com}
    \quad \texttt{alham.fikri@mbzuai.ac.ae}
    \quad \texttt{hisham.cholakkal@mbzuai.ac.ae}
}

\begin{document}

\maketitle

\begin{abstract}
In the era of foundation models, CLIP has emerged as a powerful tool for aligning text and visual modalities into a common embedding space. However, the alignment objective used to train CLIP often results in subpar visual features for fine-grained tasks. In contrast, SSL-pretrained models like DINO excel at extracting rich visual features due to their specialized training paradigm. Yet, these SSL models require an additional supervised linear probing step, which relies on fully labeled data—often expensive and difficult to obtain at scale. In this paper, we propose a label-free  prompt-tuning method that leverages the rich visual features of self-supervised learning  models (DINO) and the broad textual knowledge of large language models (LLMs) to largely enhance CLIP-based image classification performance using unlabelled images. Our approach unfolds in three key steps: (i) We generate robust textual feature embeddings that more accurately represent object classes by leveraging class-specific descriptions from LLMs, enabling more effective zero-shot classification compared to CLIP's default name-specific prompts. (ii) These textual embeddings are then used to produce pseudo-labels to train an alignment module that integrates the complementary strengths of LLM description-based textual embeddings and DINO's visual features. (iii) Finally, we prompt-tune CLIP's vision encoder through DINO-assisted supervision using the trained alignment module. This three-step process allows us to harness the best of visual and textual foundation models, resulting in a powerful and efficient approach that surpasses state-of-the-art label-free classification methods. Notably, our framework, \textbf{NoLA} (No Labels Attached), achieves an average absolute gain of 3.6\% over the state-of-the-art LaFter across 11 diverse image classification datasets. Our code and models can be found at \textcolor{red}{\url{https://github.com/fazliimam/NoLA}}.
\end{abstract}

\section{Introduction}

\begin{figure}[t]
    \centering
    \includegraphics[width=\columnwidth]{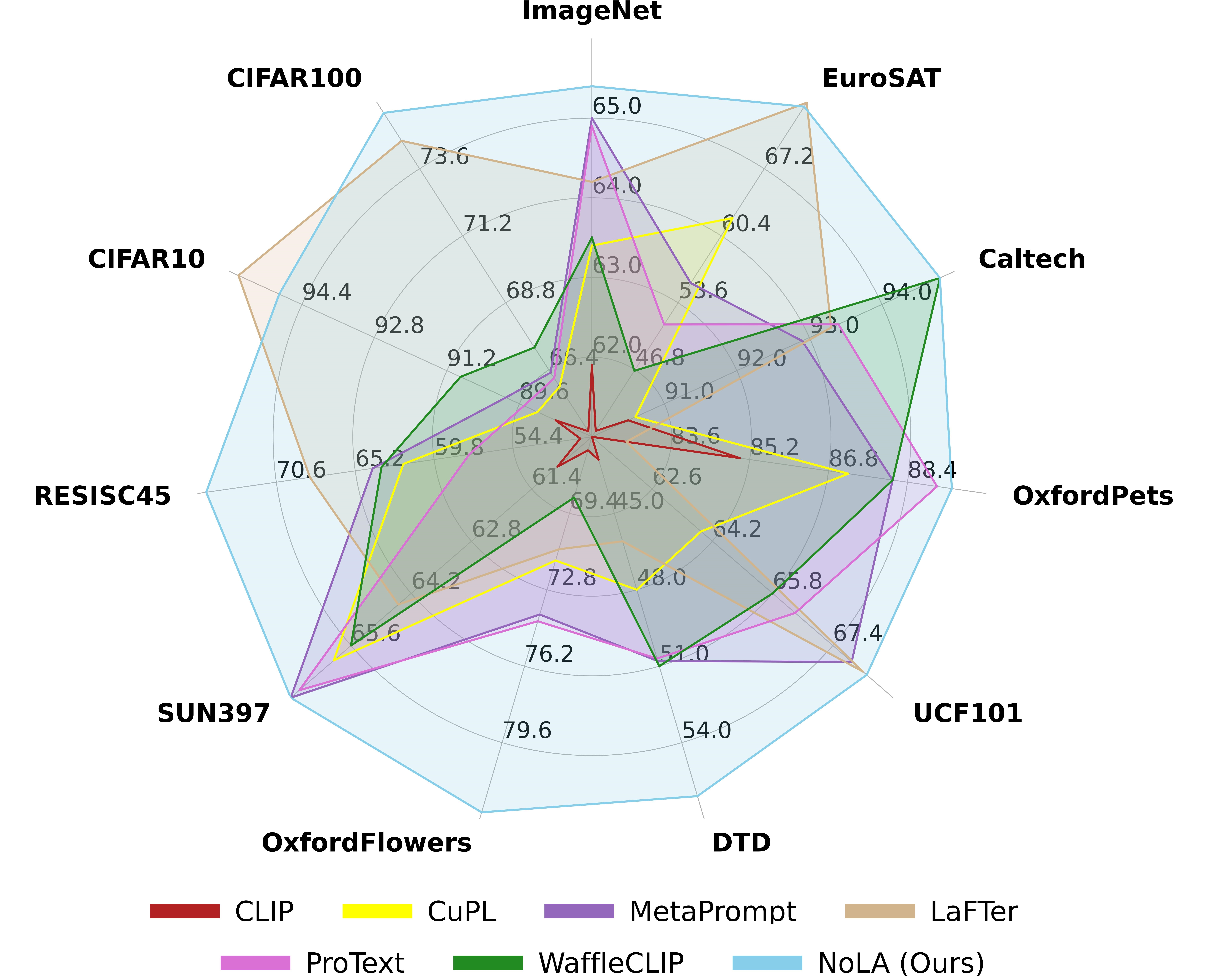}  
    \caption{Top-1 accuracy (\%) comparison with recent  label-free method on 11 diverse image classification datasets. NoLA (Ours) achieves state-of-the-art performance in 9 out of 11 datasets, outperforming  the state-of-the-art LaFter by an average absolute gain of 3.6\%.}
    \label{fig:example}
\end{figure} 
The vision-language research landscape is rapidly evolving with foundational models like CLIP \cite{radford2021learning}, ALIGN \cite{jia2021scaling}, and BLIP \cite{li2022blip} leading the charge. These models are composed of dual encoders for both text and images and map inputs to a shared embedding space, enabling the comparison of test image embeddings with text embeddings representing different classes. Among these, CLIP has gained particular attention for its ability to leverage contrastive learning on extensive image-text pairs. This innovative approach by aligning images and text representations without the need for additional data or training excels in many other computer vision tasks including medical imaging \cite{zhang2024mediclip, zhao2023clip, you2023cxr}, remote sensing \cite{qiu2022open, chen2023multiscale, yuan2023parameter, yuan2021lightweight, li2023rs, bazi2022bi}, video anomaly detection \cite{wu2024vadclip, joo2023clip} and more. Though these models offer impressive flexibility in recognizing a wide range of categories, they often require further supervised fine-tuning to match the performance of traditional methods on specific closed-set tasks. Nonetheless, the scalability and flexibility of VLMs like CLIP, ALIGN, and BLIP have significantly advanced zero-shot recognition by creating a unified representation between visual and language domains.

During pre-training, CLIP is designed to align image-text pairs within a shared feature space, enabling it to encode open-vocabulary concepts and perform effectively on zero-shot recognition tasks. CLIP includes two separate encoders—one for images and one for text. A manually crafted prompt like ``\texttt{a photo of a [CLS]}" serves as the text input during inference. The model compares the text features of different classes with the visual features, assigning the predicted label to the class with the highest similarity.
CLIP exhibits remarkable zero-shot capabilities, allowing it to make accurate predictions on tasks it has not explicitly been trained for. This is achieved through its foundation in zero-shot transfer learning, where the model leverages its extensive training on a diverse dataset of 400 million image-text pairs. By learning to associate images with their corresponding textual descriptions, CLIP can generalize its understanding to new and unseen classes. This flexibility enables it to perform various tasks, such as image classification and object detection, without requiring additional fine-tuning on labeled datasets. Consequently, CLIP represents a significant advancement in multimodal AI, effectively bridging the gap between natural language understanding and computer vision.
The joint vision-language  embedding nature of foundational VLMs such as CLIP renders such models an ideal choice for zero-shot recognition, image captioning, visual question answering, and many other tasks. However, the zero-shot recognition capability of such vision-language foundation models often falls behind the visual recognition methods trained on the target dataset. Hence, to achieve the full potential of foundation models, it is desired to adapt them towards the target dataset. Specifically, the model needs specialized adaptation to the inherent challenges in the target dataset.

On the other hand, self-supervised learning (SSL) methods have gained prominence for their ability to leverage large volumes of unlabeled data to learn meaningful representations \cite{schiappa2023self, caron2021emerging, eldele2023self, liu2023ambiguity}. Unlike traditional supervised learning, which requires extensive labeled datasets, SSL techniques generate their own supervisory signals from the data itself. SSL methods can uncover intricate patterns and relationships within the data, making them particularly valuable in scenarios where labeled data is scarce or costly to obtain. Generally, SSL methods \cite{zhu2023hnssl, park2023self, koccyiugit2023accelerating, stojnic2021self, akiva2022self} are task-agnostic, enabling a richer feature representation learning. 
Although SSL methods are initially trained using unlabeled data, their performance is often evaluated and enhanced in a fully supervised setting through techniques such as linear probing. In this approach, a simple linear classifier is trained on top of the features learned by the SSL model using a labeled dataset. This process allows researchers to assess the quality of the representations and fine-tune them for specific tasks. By leveraging supervised learning in this manner, SSL methods can effectively bridge the gap between unsupervised feature learning and practical, task-specific applications, ensuring that the learned representations are robust and useful for downstream classification and other supervised tasks.

In this paper, we introduce \textbf{\textit{NoLA (No Labels Attached)}}, an efficient method for fine-tuning CLIP to a set of finite classes without relying on any labels.  Our goal is to eliminate the need for costly image labels by employing weakly supervised fine-tuning of the CLIP model. Specifically, following \cite{pratt2023does}, we develop an enriched Class Description Embedding (CDE) classifier using descriptions generated by large language models (LLMs). This process distills the knowledge of LLMs, resulting in a more robust classifier. The LLM-enriched CDE classifier is subsequently employed to construct a DINO-based labeling (DL) network, leveraging DINO \cite{caron2021emerging}—a self-supervised learning (SSL) pre-trained vision encoder—to align with the VLM joint embedding space. Once trained, the DINO-based labeling network serves as a pseudo-labeler to learn target dataset-specific prompts, which are subsequently appended to the frozen CLIP vision encoder in a FixMatch \cite{sohn2020fixmatch} fashion.

We demonstrate the effectiveness of NoLA across 11 popular image classification datasets in a label-free evaluation, where it surpasses existing state-of-the-art methods that use LLM-generated descriptions, achieving an average gain of 3.6\% while maintaining a lightweight auto-labeling approach.

Our contributions can be summarized as follows:

\begin{itemize} 

    \item  We introduce a lightweight auto-labelled adaptation of vision language models for the classification using prompt tuning, called \textbf{No Labels Attached} \textit{NoLA}.
    \item Leveraging the rich contextual knowledge base of LLMs, we compose a class description embedding (CDE) classifier to generate pseudo labels.
    The enriched class description embedding (CDE) is used to align a pretrained SSL encoder to the VLM joint embedding space as a DINO-based labelling (DL) network. The strong visual SSL encoder, aligned towards the VLM embedding space is used as the auto-labeller towards adapting the VLM vision encoder using prompt tuning.
    \item Through an extensive evaluation on 11 widely recognized image classification datasets, we demonstrate that our method, NoLA (i) achieves an 11.91\% average improvement over 
    zero-shot CLIP and (ii) surpasses the previous state-of-the-art in a label-free setting on 9 out of the 11 datasets. Morover, our method (NoLA) achieves an average absolute
gain of 3.6\% over the state-of-the-art LaFTer, across 11 datasets. 
\end{itemize}

\section{Related Works}

\subsection{Vision-Language Models}
The vision language models (VLMs) \cite{radford2021learning, jia2021scaling, naeem2023silc, yuan2021florence, yao2021filip, yu2022coca} architecture involves three key components: firstly, utilizing a visual backbone \cite{dosovitskiy2020image} to encode visual representations; secondly, engaging a language model \cite{vaswani2017attention} to interpret the text description and generate appropriate text embeddings; and finally, consolidating a contrastive learning
objective to unify the visual representation along with language models. These VLMs are designed to attract the rich multi-modal features together for the aligned image-text pairs as well as keep distance for the un-paired image-text features which are disjoint in a unified manner. For example, CLIP \cite{radford2021learning}, ALIGN \cite{jia2021scaling}, and Florence \cite{yuan2021florence} demonstrate remarkable performance in visual representation learning and transfer learning for natural scenes.    
The resulting models act like open-vocabulary concepts and are  capable of achieving promising performance in many downstream tasks including zero-shot downstream tasks; 
such as open-vocabulary image classification \cite{ khattak2023maple, naeem2023i2mvformer}, object detection \cite{cozzolino2024raising, pan2024clip}, and segmentation \cite{liang2023open,  wysoczanska2024clip}.
Although these VLMs exhibit great performance, maintaining generalization capabilities remains a crucial challenge.

\subsection{Zero-shot Learning}

Provided the labels for the seen categories, the main objective of zero-shot learning (ZSL) is to learn a classifier that can discriminate the test samples of the unseen classes \cite{pourpanah2022review, xu2020attribute, hou2024visual}.
Recently, researchers have developed methods to enhance the zero-shot capabilities of CLIP by utilizing class-specific descriptions generated from large language models (LLMs). These methods demonstrate how a pretrained language model can create improved language prompts for open vocabulary tasks.
In these studies, hand-crafted prompts are used to query the LLM, generating visual and distinguishing attributes for the classes within the respective datasets. 

CuPL \cite{pratt2023does} was one of the earliest works to demonstrate that the prompts generated by this method can achieve superior performance on zero-shot image classification benchmarks. 
LaFTer \cite{mirza2024lafter} trains a classifier on the embeddings of generated texts for zero-shot classes, which can also be applied to image features.  
ProText \cite{khattak2024learning} utilizes a text-only supervision approach to effectively learn prompts, leveraging the capabilities of large language models (LLMs).
MetaPrompting \cite{mirza2024meta} introduces an automatic prompt generation technique to generate high-level textual information to find a way to generate diverse category-level prompts for the zero-shot classification task.
AdaptCLIPZS \cite{saha2024improved} proposes fine-grained labeling by pairing images with coarse-level descriptions which emphasizes key attributes of classes to bridge the gap between the image-level captions and generalized information of the category object, resulting in generalization to several tasks. 
WaffleCLIP \cite{roth2023waffling} proposes to introduce random descriptors for zero-shot accuracy.

\subsection{Pseudo Labelling/ Semi-Supervised Learning} 
Pseudo-labeling or semi-supervised learning is a powerful machine learning method to generate pseudo-labels of a large amount of data without requiring a large amount of labels \cite{sohn2020fixmatch, hoyer2023semivl, berthelot2019mixmatch}. It mitigates the requirement of labeled data by selecting confident ones to train models.
Inn order to boost the performance of semi-supervised learning, recent works leverage both pseudo-labeling and consistency regularization to benefit from similar predictions between the two different views of an image \cite{kurakin2020remixmatch, li2023class}. \cite{wei2023towards} propose an  adaptive consistency regularizer (ACR) method to handle the semi-supervised learning for the long-tailed classification problem. 
Further, the pseudo-labeling was extended to utilize augmentation \cite{nguyen2023boosting} and consistency regularization \cite{yan2024universal}. \cite{li2023iomatch} presents a novel open-set semi-supervised framework exploiting both inliers and outliers when they are hard to distinguish. FixMatch \cite{sohn2020fixmatch} combined consistency regularization to estimate the pseudo-label using a high-confidence prediction. It was later extended via non-parametrically predicting view assignments with support samples \cite{assran2021semi}.

\subsection{Prompt Learning} 
Over the years, machine-learning approaches generally focused on fully supervised learning, which employs task-specific models that are exclusively trained on instances with labels relevant to the target task \cite{krizhevsky2012imagenet, alom2018history}. In the recent era of foundation models, the learning paradigms have undergone considerable modernization and are moving away from fully supervised learning to a \textit{pre-training} and \textit{fine-tuning} learning frameworks for the downstream tasks \cite{zhou2022conditional, zhou2022learning, gao2024clip}. These approaches leverage the models to acquire generalized feature learning during the pre-training and do not require exclusively adapting the model to downstream tasks. On the contrary, researchers are redesigning the inputs using prompts to revamp the downstream task ensuring that it corresponds with the original pre-training task \cite{lester2021power, lu2022prompt}. Prompt learning has shown the promising potential to minimize semantic discrepancies and bridge the gap between pre-training and fine-tuning to overcome the issues related to the overfitting problem \cite{lu2022prompt, liu2024visual, khattak2023maple, lee2023multimodal}.
CoOP\cite{zhou2022learning} proposes that the prompt vectors by employing the cross-entropy loss can reduce the prediction error. 
CoCoOp \cite{zhou2022conditional} introduce which generates image-adaptive prompts resulting in enhanced generalization to the distribution shifts. The unsupervised prompt learning (UPL) approach  \cite{huang2022unsupervised}  avoids the prompt engineering. Whereas, TPT optimizes the prompt by minimizing the entropy with confidence selection \cite{shu2022test} by introducing a test-time prompt learning framework.

\section{Methodology}

We first provide an overview of CLIP and prompt learning. We then introduce our framework, No Labels Attached (NoLA) tuning and provide a detailed explanation of how we apply our method, combining the strengths of VLMs and pre-trained self-supervised learned vision backbones, for improved performance.

\begin{figure*}[t]
\centering
\includegraphics[width=1\textwidth]{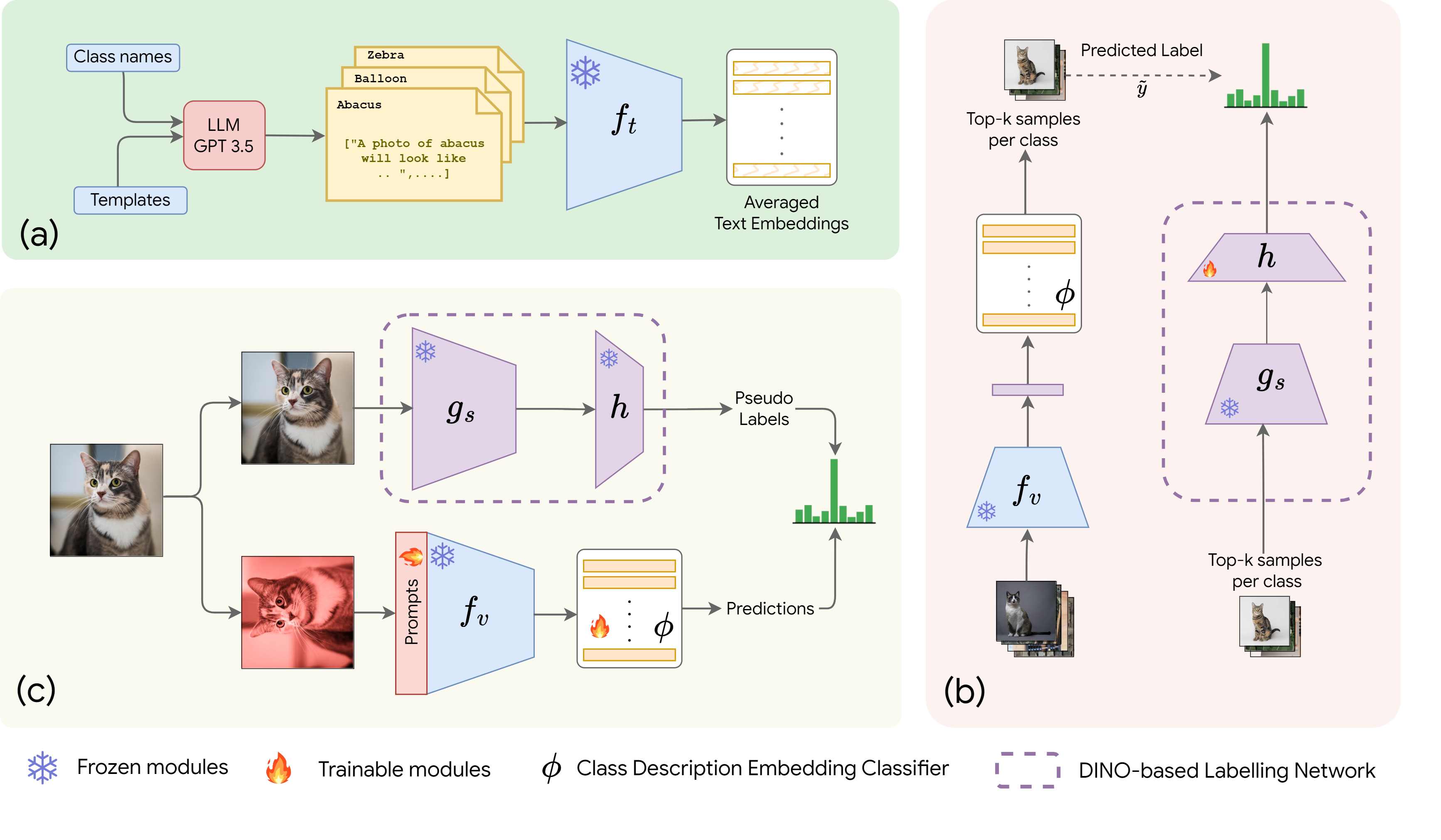} 
\caption{\textbf{Overview of proposed \textit{NoLA} (No Labels Attached) method,} \texttt{(a)} A set of templates and the class names are fed through an LLM to generate context-enriched text descriptions per class. The description embeddings obtained from the CLIP text encoder \bm{$f_{t}$} are averaged to compose the class description-based embedding (CDE) classifier \bm{$\phi$}. \texttt{(b)} Zero-shot inference is obtained for the training set by using the CLIP vision encoder \bm{$f_{v}$} and the CDE classifier. From the predictions, top-k confident training samples are selected to train the alignment module $h$ which utilizes a self-supervised learned (SSL) \bm{$g_{s}$} backbone (DINO). \texttt{(c)} The DINO-based labelling network consisting of the alignment module $h$ is then used to generate pseudo labels and learn dataset specific visual prompts which are prepended to the frozen CLIP vision encoder.}
\label{NOLA_arc}
\end{figure*}

\subsection{Preliminaries}
\label{subsec:preliminaries}

\noindent\textbf{\textit{Contrastive Language-Image Pre-training (CLIP):}}
CLIP comprises two parallel encoders, mapping the visual and textual inputs into feature vectors in the joint embedding space. Here, the CLIP image and text encoders are denoted by $\mathcal{F}_v$ and $\mathcal{F}_t$ respectively, and their pre-trained parameters are represented by ${\theta}_{\mathtt{CLIP}} = \{\theta_{v}, \theta_{t} \}$ respectively. An input image ${I}$ is converted to $M$ patches, which are projected to produce patch tokens, and a class token ${\textsc{cls}}$ is prepended to it, resulting in $\bm{{X}}_{0}= \{{\textsc{cls}}, \bm{p}_{1}, \bm{p}_{2}, \hdots, \bm{p}_{M}\}$ where $e_i$ is the embedding of the $i^{th}$ patch. The image encoder $\mathcal{F}_v$ encodes the input patches via transformer blocks to produce a latent visual feature representation $\bm{{f}_v} =\mathcal{F}_v(\bm{{X}}_{0}; \theta_{v})$. The corresponding class label ${y}$ is embedded within a text template or description, such as `\texttt{a photo of a}  $<{\textsc{cls}}>$', which is tokenized to form $\bm{{Y}}_{0}$. 
The text encoder ${\mathcal{F}_t}$ encodes $\bm{{Y}}_{0}$ through transformer blocks to compute the latent textual feature as $\bm{{f}_t} = \mathcal{F}_{t}(\bm{{Y}}_{0}; \theta_{t})$. At zero-shot inference, the similarity of each text feature with class labels $y = \{1, 2, \cdots, C\}$ is computed with that of the image feature as $\bm{s_i} = \mathtt{sim}(\bm{{f}_{t_{i}}} \cdot \bm{{f}_v})$, where $\mathtt{sim}(.)$ denotes the cosine similarity, $\bm{s_i}$ denotes the similarity score of $i^{th}$ class with the text feature $\bm{{f}_{t_{i}}}$. The prediction probability $p(y_i|{X})$ on ${X}$ can be defined as:

\begin{equation}
     p(y_i|{X}) = \frac{\mathtt{exp}(\mathtt{sim}(\bm{{f}_t} \cdot \bm{{f}_v})\tau)}{\sum_{i=1}^{C}\mathtt{exp}(\mathtt{sim}(\bm{{f}_t} \cdot \bm{{f}_v})\tau)},
\end{equation}
  where $\tau$ is the temperature of the softmax function.

\noindent\textbf{\textit{{Prompt Learning:}}}
CLIP contains a plethora of knowledge leveraged from training on millions of noisy image-text pairs. 
To effectively extract the rich features learned by the CLIP model, 
recent approaches \cite{zhou2022learning, zhou2022conditional, khattak2023maple, lu2022prompt} append extra learnable prompts while keeping image and text encoders frozen. These prompts modify the context of the model input without distorting the pre-trained CLIP features. We  prepend visual prompts on the vision encoder of CLIP, denoted by ${\theta}_{\mathtt{P}}$.

\subsection{Motivation}
\label{subsec:ALP-RS}
While CLIP's native zero-shot classification capability shows promising results without requiring training data, it is usually surpassed by networks trained on data specific to the target domain. Existing methods \cite{zhou2022learning, zhou2022conditional, saha2024improved} narrow this performance gap through fine-tuning the zero-shot classifier in a few-shot setting. 
While improving accuracy they also incur additional costs associated with curating and annotating training data. 

Additionally, the features extracted using the pretrained CLIP vision encoder are often sub-optimal for many fine-grained visual recognition tasks, as they do not effectively discriminate between the distinguishing characteristics of similar-looking image categories. While there are models that demonstrate superior feature extraction capabilities—specifically, those trained in a self-supervised manner like DINO \cite{caron2021emerging}, SimCLR\cite{chen2020simpleframeworkcontrastivelearning} —these models often require labeled datasets to learn a linear probing adapter for downstream datasets. We aim to leverage the rich visual features learned in self-supervised models, particularly DINO pretrained on ImageNet \cite{deng2009imagenet}, to adapt large VLMs such as CLIP by utilizing only unlabeled training images. DINO pre-trained on ImageNet is particularly advantageous because it captures a wide range of visual features across diverse categories, enabling more effective adaptation to fine-grained tasks.

\subsection{Overview}
An overview of the proposed approach is shown in Figure \ref{NOLA_arc}. which comprises the following three components (i) A class description-based embedding (CDE) classifier that builds textual embeddings using class descriptions prompted through an LLM, enriched by their vast knowledge base (As shown in Figure  \ref{NOLA_arc}.-(a)). (ii) We then use a stronger SSL pre-trained visual backbone such as DINO and align it to the joint embedding space of the VLM to be used as the auto-labelling network (As shown in Figure \ref{NOLA_arc}.-(b)). (iii) This stronger SSL pre-trained visual backbone is then used for the DINO-assisted prompt learning in the vision encoder (As shown in Figure \ref{NOLA_arc}.-(c)). Next, we provide detailed explanation of these three key components.

\subsubsection{Class Description based Embedding (CDE) classifier:} 

The generic vision-language model setting uses their text encoder to build the classifier to classify the image embedding, given the class names of the target dataset. We compose the classifier by generating finer descriptions catered towards the target dataset by prompting an LLM model, a technique adopted from \cite{pratt2023does}. We prompt the LLM with the class names and $N$ template questions, specific to the target dataset as shown in Figure \ref{NOLA_arc}.-(a). This generates $K$ descriptions $\bm{\omega^C}$ for each class $C$, enriched by the domain knowledge of the LLM, giving $K \times C$, class descriptions. 
The class description based embedding classifier $\bm{\phi} \in \mathbb{R}^{C \times d}$, where $d$ is the embedding dimension, can be formulated as follows:
\begin{align}
\begin{split}
\label{eq:CDTE}
    \bm{\phi_{C}} &= \frac{1}{K}\sum_{i=1}^{K}{\mathcal{F}_{t}(\bm{\omega^C}_i; \theta_{t})} \\
    \bm{\phi} &= \mathtt{Concat} [\bm{\phi_{1}}, \bm{\phi_{2}}, \hdots , \bm{\phi_{C}}].
\end{split}
\end{align}

\subsubsection{DINO-based Labelling (DL) Network:}

We seek to improve the visual embedding by utilizing a strong self-supervised pre-trained visual backbone $\bm{g_{s}}$. To this end, we make use of the LLM-enriched CDE classifier, to align a self-supervised learning (SSL) pre-trained vision backbone to the joint embedding space of the VLM. \par

In order to obtain the text-aligned visual embedding $\bm{h}$ for the target dataset, we first input the target image to a CLIP visual encoder  $\bm{{f}_v}$ to output visual features. These features are further fed to the CDE classifier ($\bm{\phi_{C}}$) to obtain the top-k samples for each class $\bm{C}$, here \textit{k} is proportional to the number of training images and the number of classes in the respective target dataset. It is only fair to select a higher value for \textit{k} in datasets with more samples per class, and a lower value otherwise. Since our method is entirely label-free, we do not use information about the number of samples per class. Instead, we determine k using the data available to us: the number of training images and the number of classes. First, we calculate the average number of images per class by dividing the total number of training images by the number of classes. To account for inherent class imbalance that will be present in most datasets, we then select 20\% of this average. The choice of 20\% as the optimal percentage is justified by empirical analysis, which we present in the supplementary material. Additionally, if the calculated \textit{k} is less than 16, we set \textit{k} to 16, and if it exceeds 512, we cap \textit{k} at 512.
Later, the alignment module $\bm{h}$ is then optimized utilizing smoothed cross-entropy loss function \cite{szegedy2016rethinking}, to obtain a DINO-based labelling (DL) network (comprising of $\bm{{g}_s}$ and $\bm{h}$), using the top-k samples per class, 
where all other components are frozen (Fig~\ref{NOLA_arc}-(b)).

\subsubsection{DINO-assisted prompt learning:}

In order to adapt the vision encoder of the VLM, we set up learnable visual prompt tokens ${\theta}_{\mathtt{P}}$ to the vision encoder.
Specifically, we append learnable $V$ visual prompts with the visual input tokens. The image encoder processes the input to generate a prompted visual feature representation denoted as $\bm{{f}^p_v}$ can be represented as:
\begin{equation}
    \bm{{f}^p_v} = \mathcal{F}_{v}(\bm{{X}}_{0}; \theta_{v}, {\theta}_{\mathtt{P}}).
\end{equation}
This facilitates a lightweight adaptation of the vision encoder as opposed to fine-tuning the vision encoder.
To do so, motivated by Fixmatch \cite{sohn2020fixmatch}, we generate two separate views for each target input i.e., weak transformation as identity ($\bm{I_0}$) and strong augmentation ($\bm{I_s}$). This approach is effective because weak augmentation tends to preserve the intrinsic characteristics of the input data, facilitating the creation of pseudo-labels that are more reliable and, strong augmentation introduces perturbations that encourage the model to learn robust and invariant features, thus improving its generalization capability. Thus, striking a balance between ensuring label quality and improving the model's capacity to generalize to previously unseen data.

We now use the DL Network --a vision encoder with finer visual cues, aligned to the VLM embedding space-- as an auto-labeller to train the visual prompts and the CDE classifier for the target dataset, through the training objective given in Eq.~\ref{eq:ALP-RS}. 
\begin{align}
\label{eq:ALP-RS}
    \min_{{\theta}_{\mathtt{P}}, {\phi}} {\mathcal{L_\mathtt{SCE}} \bigg( \bm{\phi} \big( \mathcal{F}_{v}(\bm{{X}}_{s}; \theta_{v}, {\theta}_{\mathtt{P}}) \big) , \bm{h} \big( \bm{g_{s}}(\bm{{X}}_{0}; \theta_{g}) \big) \bigg)},
\end{align}
where $SCE$ represents the smoothed cross-entropy loss function \cite{szegedy2016rethinking}, the  $\theta_{g}$ denotes the pre-trained parameters of $\bm{g_{s}}$. The  $\bm{X_s}$ and $\bm{X_0}$ are the patchified input of strong and weak augmented images $\bm{I_s}$ and $\bm{I_0}$, respectively.

Through lightweight auto-labelled prompt tuning setting (Figure ~\ref{NOLA_arc}.-\texttt{(c)}), we harmonically combine the domain knowledge distilled from the LLM using the CDE classifier and stronger visual cues from a pre-trained SSL encoder towards better performance for the label-free classification task.

\section{Experiments}

\input{tables/main_table_32}

\subsection{Datasets}
We extensively evaluate our approach across 11 diverse datasets, each representing distinct domains. Among these, four datasets—ImageNet \cite{deng2009imagenet}, CIFAR-10/100 \cite{krizhevsky2009learning}, and Caltech-101 \cite{fei2006one}—focus on common natural categories. EuroSAT \cite{helber2019eurosat} and RESISC45 \cite{cheng2017remote}, each containing 10 and 45 classes respectively, provide satellite imagery for geographical and environmental analysis. The UCF-101 dataset \cite{soomro2012ucf101} is used for action recognition, while SUN-397 \cite{xiao2016sun} offers images from 397 naturally occurring scenes. Flowers-102 \cite{nilsback2008automated} is a fine-grained classification dataset containing 102 different categories of flowers. The Describable Textures Dataset (DTD) \cite{cimpoi2014describing} comprises 47 categories of images, designed to study texture perception through various describable attributes. Lastly, the Oxford Pets \cite{parkhi2012cats} dataset features 37 categories of pet images, covering a range of cat and dog breeds. For all the datasets, we either use the splits provided by the author or, if unavailable, we use the split provided by \cite{zhou2022learning}.

\subsection{Implementation Details}
As discussed earlier, our pipeline includes three main stages. In the first step, we utilize the descriptions obtained from an LLM i.e., GPT3.5, we use the descriptions dataset obtained by \cite{pratt2023does} in which they prompt with the class names of the target dataset and $N$ dataset specific questions to generate $K$ class-specific descriptions. For the obtained descriptions, we also add dataset-specific prompt templates provided by \cite{radford2021learning}.

In the second stage, we build the DINO-based labelling (DL) network, using a self-supervised vision encoder aligned to the VLM joint embedding space. We keep the DINO ViT-B/16 \cite{caron2021emerging}, Imagenet pre-trained backbone, $\bm{g_s}$ (in Fig~\ref{NOLA_arc} \texttt{(b)}) frozen and train  alignment module $\bm{h}$ on the target dataset.  The choice of k value in the top-k samples to be selected is different for each dataset and it is propotional to the number of training images and number of categories in the dataset. The specifications, training details of the alignment module $\bm{h}$ and reasoning behind k value selection are mentioned in the supplementary material. \par

In the final DINO-assisted prompt learning stage, we include learnable prompts in the vision encoder of the VLM. The DINO-assisted prompt learning is performed using the AdamW \cite{kingma2014adam} optimizer with a learning rate of $2e^{-3}$ and a batch size of $512$. To obtain the augmented view of the image, we employ augmentations from SimSiam \cite{chen2021exploring}: Gaussian blur, random resized crop, random horizontal flip, color jitter, random scaling, and, random perspective. All experiments are conducted using a single Nvidia A100 GPU.

\subsection{Results and discussion}
We evaluate the image classification performance of NoLA across the 11 datasets presented in Table~\ref{tab:vitb32_main_results} using ViT-B/32 \cite{dosovitskiy2020image} CLIP variant. We compare the performace our method with six label-free methods CuPL \cite{pratt2023does}, MetaPrompt \cite{mirza2024meta}, LaFTer \cite{mirza2024lafter}, ProText \cite{khattak2024learning}, WaffleCLIP \cite{roth2023waffling}, and also CLIP zero-shot performance. We also conduct a quantitative analysis by comparing our method with CoOp \cite{zhou2022learning}, a dew-shot method.
As seen in Figure \ref{tab:vitb32_main_results}, our approach demonstrates state-of-the-art performance in 9 out of 11 datasets when compared with label-free methods and even, outperforms few-shots methods on certain datasets.

\begin{figure}[h]
    \centering
    \includegraphics[width=0.95\linewidth]{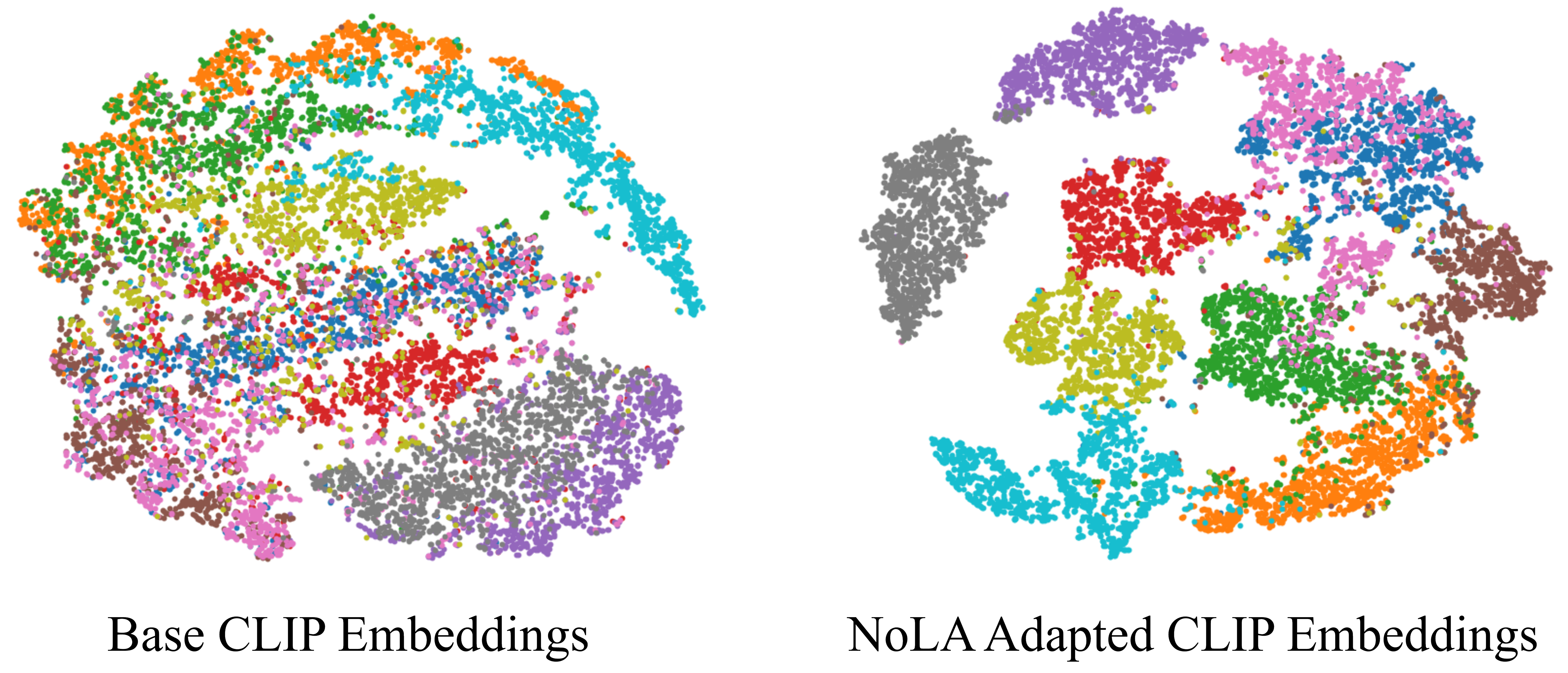}  
    \caption{TSNE projections comparison of EuroSAT embeddings obtained from the base CLIP ViT-B/32 (left) and CLIP ViT-B/32 adapted with our NoLA framework (right).}
    \label{fig:tsne_eurosat}
\end{figure}
In addition, in Figure \ref{fig:tsne_eurosat}, we compare visualizations of embeddings from base CLIP and our method. It is evident that our method produces better pronounced, discriminative clustered embeddings as same class features are closer to each other, while different class features are far apart.

\input{tables/main_table_16}

\input{tables/ablation_table_ind_comp}

\subsection{Ablation Study}
We conduct an ablative study for proposed method NoLA. The experiments are conducted across six datasets namely, EuroSAT, Caltech101, OxfordPets, OxfordFlowers, SUN397, and CIFAR100 for all ablation experiments. \par

\subsubsection{Using CLIP ViT-B/16 variant:} Table~\ref{tab:vitb16_main_results} demonstrates that our methodology maintains strong performance with the CLIP ViT-B/16 variant. Additionally, our approach achieves results comparable to AdaptCLIPZS \cite{saha2024improved}, a few-shot method utilizing 16 shots.

\subsubsection{The different stages of NoLA:} Table~\ref{tab:ablation1} summarizes the contribution of each stage-wise component in our proposed method using CLIP ViT-B/32 variant.
As observed, the class-specific LLM knowledge distilled through the CDE provides a significant boost in performance with an increase of $4.1\%$ against the zero-shot accuracy of CLIP. Next, the performance of the trained DL network shows a further improvement of $1.9\%$. This validates the importance of combining the strengths of VLMs and enriched visual features from a stronger visual backbone. A further boost of $6.6\%$ in the performance is obtained through the DINO-assisted prompt learning method, adapting the vision encoder of the VLM using prompt learning, thus showcasing the significance of each stage.

\subsubsection{Ablation of design choices:} Table~\ref{tab:ablation_combined} illustrates two ablations where in ablation 1 shows the averaged Top 1 \% accuracy obtained when the DINO vision encoder is replaced with CLIP vision encoder and ablation2 shows the averaged Top 1 \% accuracy obtained when trained DL network is completely replaced with CDE classifier. While both these setups are able to get much better performance in accuracy when compared to CLIP zero-shot, the incorporation of a rich feature extractor like DINO and enriched class embeddings brings in a more nuanced understanding of visual features, leading to further improvements in NoLA's discriminative ability as shown in the overall accuracy.
\input{tables/ablation_combine}

\section{Conclusion}

In this work, we propose a label-free lightweight prompt tuning for vision language models. Particularly, we leverage knowledge from the Large Language Model (LLM) to build a class description embedding (CDE) classifier and use pseudo-labels from the CDE classifier to align an SSL pre-trained vision encoder, DINO, to the vision-language joint embedding space, to build the DINO-based Labelling (DL) network. Finally, we employ our trained DL network as an auto-labeller to adapt the vision-language vision encoder through prompt tuning. 
We perform extensive experiments over 11 popular image classification datasets and our study reveals that our framework, NoLA, performs favorably compared to existing VLMs-based state-of-the-art methods.

\section{
Supplementary Material: CLIP meets DINO for Tuning Zero-Shot Classifier using 
\\
Unlabeled Image Collections
 }

\noindent In this supplementary, we provide,
\begin{itemize}
    \item Ablation of trainable components in NoLA    
    \item Ablation of using GPT-4o descriptions
    \item Analysis on $k$ (number of confident pseudo-labels per class) selection
    \item Implementation of class description embedding (CDE) classifier
    \item Additional implementation details
\end{itemize}
All ablations and experiments in this supplementary material are conducted using the ViT-B/32 CLIP variant unless specified otherwise.

\section{Ablation of trainable components in NoLA}
We summarize the findings in Table~\ref{tab:suppl_3} to evaluate the impact of different trainable components when adapting the vision encoder through DINO-assisted prompt learning. We observe that the combined training of both the visual prompts and the learnable CDE (as shown in Fig 2-(c) in the main paper) yields a reasonable improvement compared to making only one of these components trainable (Settings 2 or 3). This enhancement can be attributed to the synergistic benefits of visual adaptation via prompts and the domain knowledge captured by the LLM-derived CDE classifier.

\input{tables/suppl_3}

\section{Ablation of using GPT-4o descriptions}
While we utlize the descriptions dataset obtained from \cite{pratt2023does}, which is generated using GPT3.5, we also experiment the performance of our framework when paired with richer descriptions generated from GPT-4o. The prompts used to generate the descriptions are provided in the Appendix. The findings of this ablation is presented in Table \ref{tab:gpt4omini}, where we compared Top-1 accuracy obtained for the six datasets which were used in the ablations. This comparison allows us to assess the impact of more detailed and contextually rich descriptions on the overall performance of our framework. Notably, GPT-4o descriptions achieve an average accuracy that is 0.64 higher than with GPT-3.5.
\input{tables/ablation_table_llm}

\section{Analysis on $k$ selection}

The alignment module $h$ within the DL network is trained on samples selected using the CDE classifier, making the number of confident pseudo-labels per class, $k$, crucial for training. \cite{pantazis2022svl, huang2022unsupervised} argued that choosing $k$ as 16 is optimal for any dataset. However, we hypothesize, for a dataset which has a higher number of images per class, it's reasonable to select a higher number of pseudo-labels. To explore this, we analyzed the impact of different values of $k$ on DL network performance.

Since our method is entirely label-free, we cannot directly use the information on the number of training images available for each class. Instead, we estimate the average number of images per class by dividing the total number of training images by the number of classes. However, this estimate may not accurately represent the true distribution due to the imbalanced nature of datasets in the wild. To account for the long-tailed distributions, we select only a proportion of this estimated value. We experiment with different proportions of the average number of images per class to determine the optimal proportion. Specifically, we test selecting between 10\% and 30\% with 5\% increments.

\begin{figure}[!htb]
    \centering
    \begin{subfigure}[b]{0.23\textwidth}
        \centering
        \includegraphics[width=\textwidth]{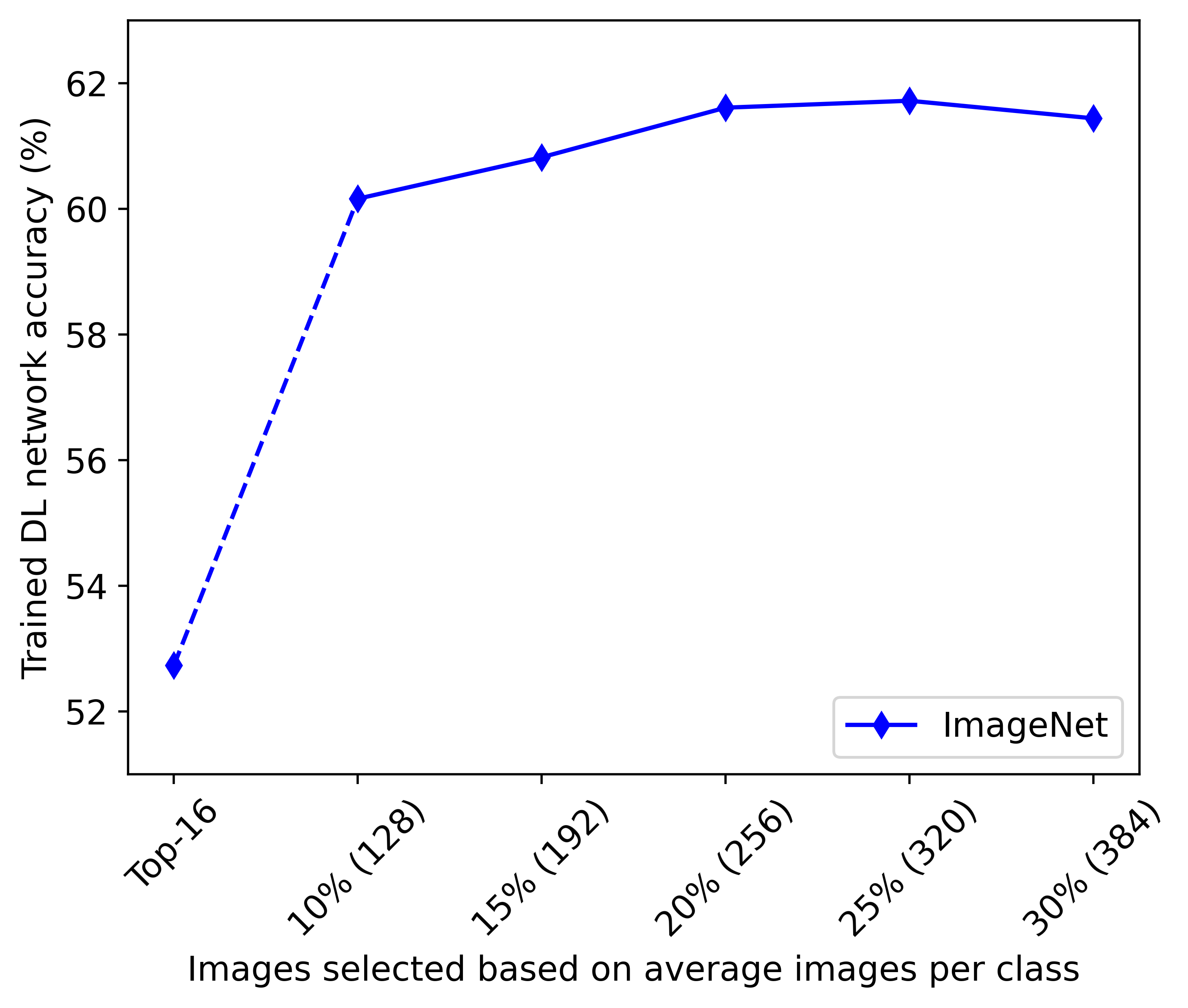}
        \label{fig:image1}
    \end{subfigure}
    \hspace{-0.4cm}  
    \begin{subfigure}[b]{0.23\textwidth}
        \centering
        \includegraphics[width=\textwidth]{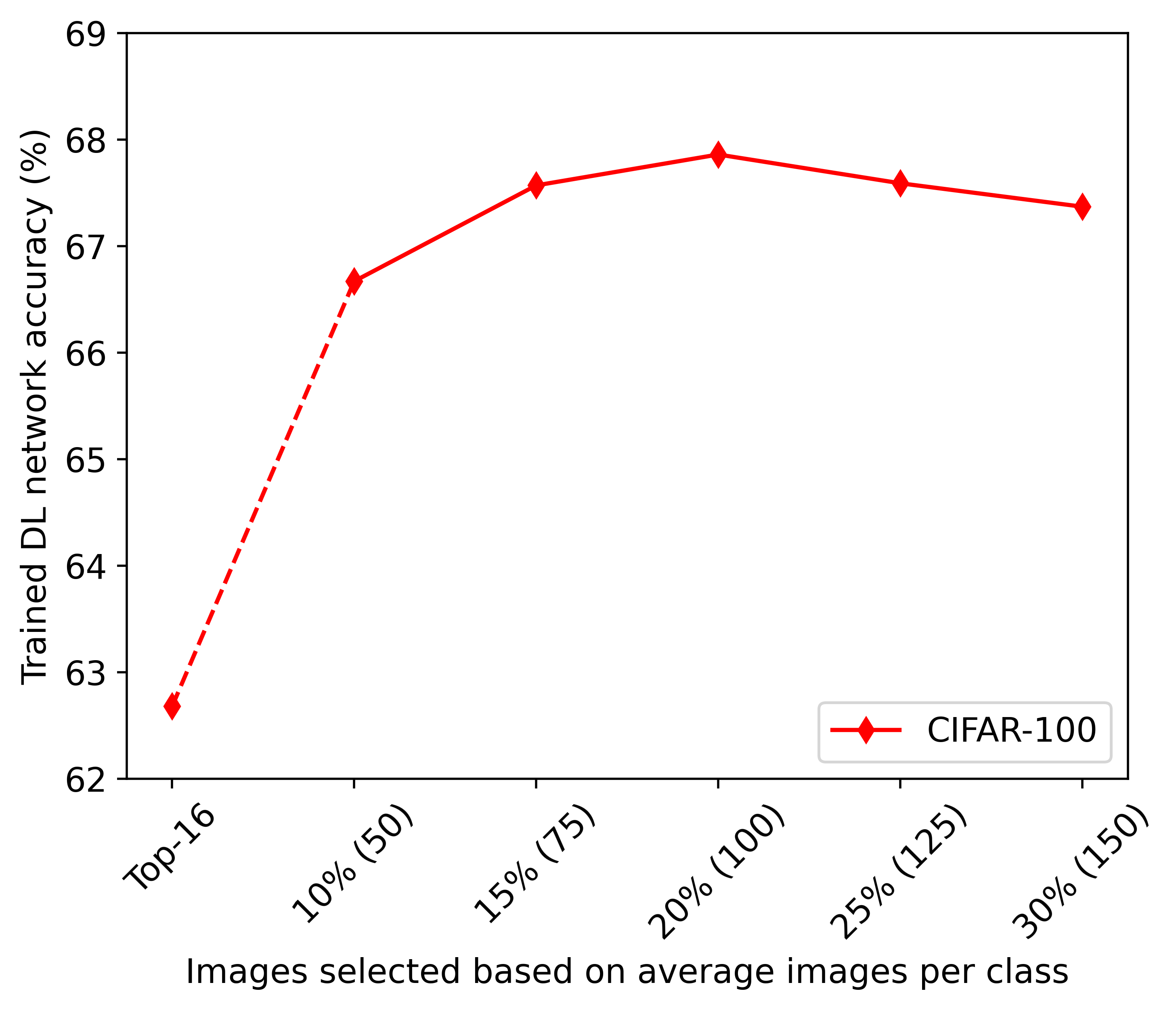}
        \label{fig:image2}
    \end{subfigure}
    
    \vspace{0.1cm}  

    \begin{subfigure}[b]{0.23\textwidth}
        \centering
        \includegraphics[width=\textwidth]{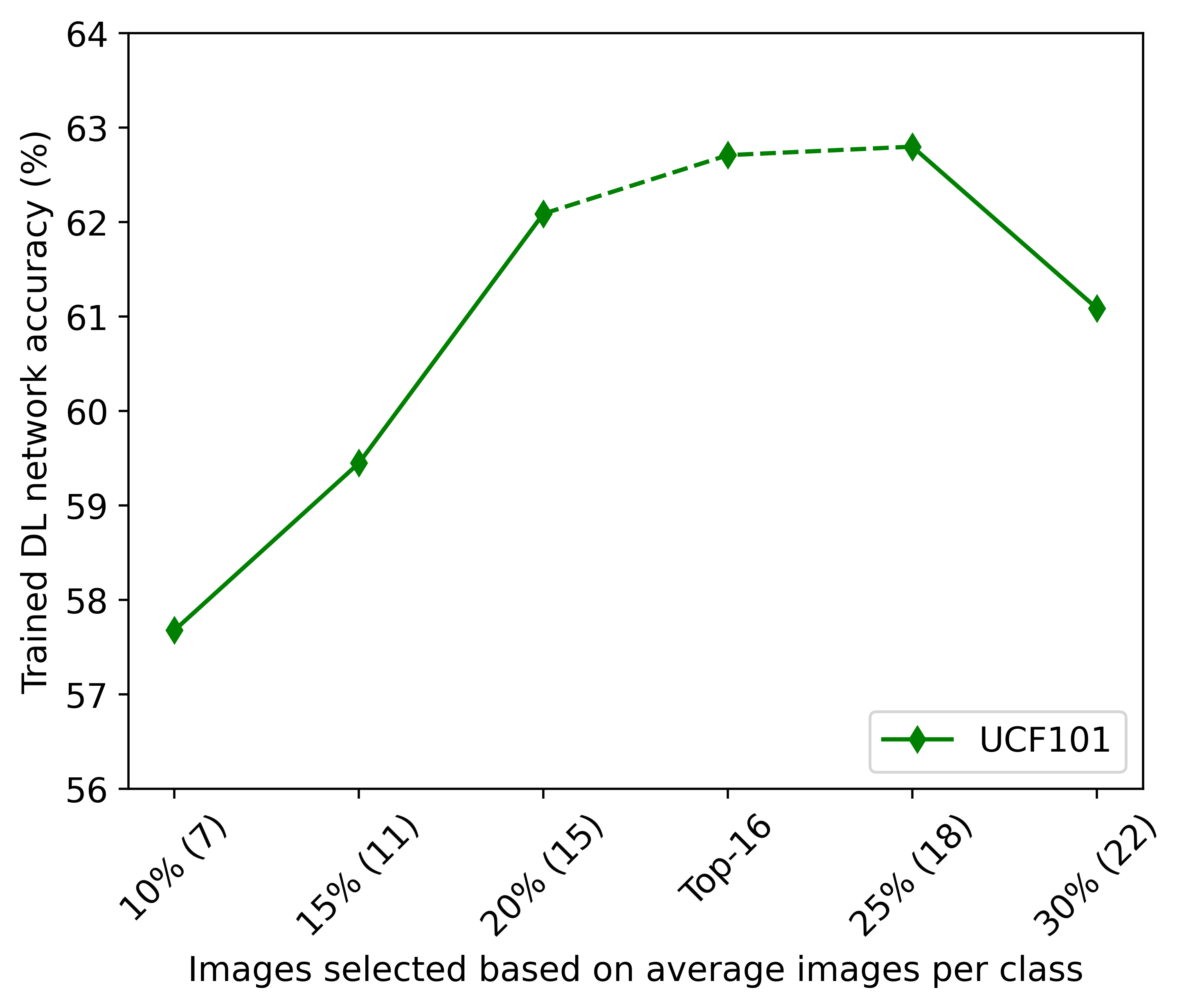}
        \label{fig:image3}
    \end{subfigure}
    \hspace{-0.4cm}  
    \begin{subfigure}[b]{0.23\textwidth}
        \centering
        \includegraphics[width=\textwidth]{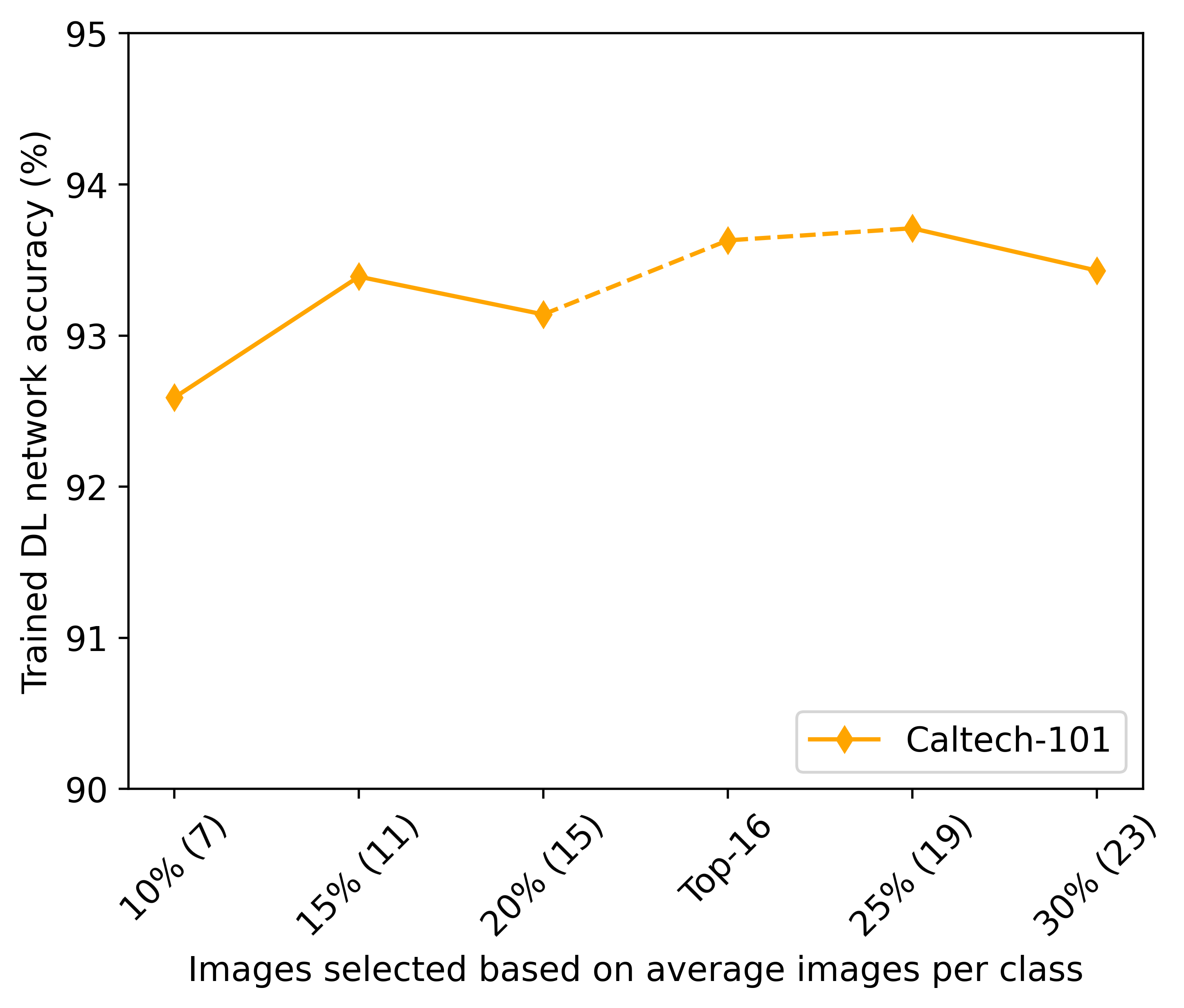}
        \label{fig:image4}
    \end{subfigure}
    
    \caption{\textbf{Top-1 Accuracy of trained DL network with different values for $k$.} The top row shows the performance of different values for $k$ in large datasets, namely, ImageNet (top-left) and CIFAR-100 (top-right). The bottom row shows the performance of different values for $k$ in small datasets, namely, UCF101 (bottom-left) and Caltech101 (bottom-right). The value inside the parentheses on the x-axis represents the number of pseudo labels selected according to the specified percentage.}
    \label{fig:topk}
\end{figure}

As shown in Figure \ref{fig:topk} (bottom), for smaller datasets, setting \( k \) to 16 yields better performance \cite{pantazis2022svl, huang2022unsupervised}. In contrast, for larger datasets, we find that \( k \) set to 16 is suboptimal. Empirically, setting \( k \) to around 20\% of the average number of images per class achieves better accuracy (see Figure \ref{fig:topk} - top).  Thus, we adopt the following strategy: if 20\% of the average number of images per class is less than 16, we select 16 confident samples. Otherwise, we select 20\% of the confident samples, with a cap of 512 if the number exceeds this limit.

\section{Implementation of class description embedding (CDE) classifier}
We develop a class description embedding (CDE) classifier, enhanced by descriptions generated from the extensive knowledge base of large language models (LLMs), a technique we adopt from \cite{pratt2023does}. 
For a dataset with $N$ classes, for any given class $n \in \{1, \cdots ,N\}$  we create textual descriptions $\{T_{n,m}\}_{m=1}^M$, where $T_{n,m}$ denotes the $m^{\text{th}}$ description of the $n^{\text{th}}$ class. These descriptions capture a wide range of semantic information, enriching the classifier's ability to distinguish between different classes. 

For each description $T_{n,m}$, we find the corresponding text embedding $\phi_{n,m}=\mathcal{F}_t(T_{n,m})$ and with this an average text embedding is computed for each class $n$ as, 
\begin{equation}
    \bm{\phi_{n}} = \frac{1}{M}\sum_{i=1}^{M}{\phi_{n,m}}
\end{equation}

Finally the CDE classifier $\phi$ is constructed by concatinating the classwise average text embeddings as, $\bm{\phi}= \mathtt{concat} [\bm{\phi_{1}}, \bm{\phi_{2}}, \hdots, \bm{\phi_{N}}]$.

\section{Additional implementation details}
Table \ref{tab:suppl_1} provides detailed hyperparameters for training the alignment module $\bm{h}$ (as shown in Fig 2-(b) of the main paper) within the DINO-based labeling (DL) network. Similarly, Table \ref{tab:suppl_2} presents the hyperparameters used for the DINO-assisted prompt learning stage (as shown in Fig 2-(c) of the main paper).

\input{tables/suppl_1}
\input{tables/suppl_2}

\bibliography{main}
\newpage
\appendix

\onecolumn
\begin{center}
\large{\sc Appendix: \emph{CLIP meets DINO for Tuning Zero-Shot Classifier using Unlabeled Image Collections}}\\
\end{center}

Below we have listed the prompts used to generate descriptions for various datasets from GPT-4o.

\begin{table}[!htb]
\begin{tabular}{l}
\textbf{\Large ImageNet, Caltech101, CIFAR10, CIFAR100, SUN397} \\ \midrule
    \textbf{Prompts} \\ \midrule
    Describe what a(n) \{\} looks like \\
    What does a(n) \{\} look like? \\
    What characteristics can be used to differentiate a(n) \{\} from others based on just a photo? \\
    Describe an image from the internet of a(n) \{\} \\
    A caption of an image of a(n) \{\}: \\
    List how one can recognize the a(n) \{\} within an image. \\
    List the distinguishing features of the a(n) \{\}. \\
    List the visual cues that help in identifying the a(n) \{\}. \\
    List the visual characteristics that make the a(n) \{\} easily identifiable. \\
    List how one can identify the a(n) \{\} based on visual cues. \\
\end{tabular}

\label{tab:prompts_1}
\end{table}

\begin{table}[!htb]
\begin{tabular}{l}
\textbf{\Large EuroSAT, RESISC45} \\ \midrule
    \textbf{Prompts} \\ \midrule
    Describe a satellite photo of a(n) \{\} \\
    Describe a(n) \{\} as it would appear in an aerial image \\
    How can you identify a(n) \{\} in an aerial photo? \\
    Describe the satellite photo of a(n) \{\} \\
    Describe an aerial photo of a(n) \{\} \\
    List how one can recognize the a(n) \{\} within an aerial image. \\
    List the distinguishing features of the a(n) \{\} in a satellite photo. \\
    List the visual cues that help in identifying the a(n) \{\} in an aerial image. \\
    List the visual characteristics that make the a(n) \{\} easily identifiable in a satellite image. \\
    List how one can identify the a(n) \{\} based on visual cues in an aerial photo. \\
\end{tabular}

\label{tab:prompts_2}
\end{table}

\begin{table}[!htb]
\begin{tabular}{l}
\textbf{\Large Flowers102} \\ \midrule
    \textbf{Prompts} \\ \midrule
    Describe how to identify a(n) \{\}, a type of flower. \\
    Describe a photo of a(n) \{\}, a type of flower. \\
    What does a(n) \{\} flower look like? \\
    List the distinguishing features of a(n) \{\} flower. \\
    How can you recognize a(n) \{\} flower in a photo? \\
    Describe the visual characteristics of a(n) \{\}, a flower of the \{\} category. \\
    What visual cues help identify a(n) \{\} flower? \\
\end{tabular}
\label{tab:flower_prompts}
\end{table}

\begin{table}[!htb]
\begin{tabular}{l}
\textbf{\Large Oxford-Pets} \\ \midrule
    \textbf{Prompts} \\ \midrule
    Describe what a pet a(n) \{\}\textbf{ }looks like. \\
    Describe a photo of a(n) \{\}, a type of pet. \\
    Visually describe a(n) \{\}, a type of pet. \\
    List the distinguishing features of a(n) \{\}\textbf{ }pet. \\
    How can you recognize a(n) \{\}\textbf{ }pet in a photo? \\
    Describe the visual characteristics of a pet a(n) \{\}.\textbf{ }\\
    What visual cues help identify a(n) \{\}\textbf{ }pet? \\
    Describe how to identify a pet a(n) \{\}\textbf{ }in an image. \\
\end{tabular}
\label{tab:pet_prompts}
\end{table}

\end{document}

%% file: tables/main_table_32.tex
\begin{table*}[]
\centering
\renewcommand{\arraystretch}{1.3} 
\setlength{\tabcolsep}{3pt}
\scalebox{1.0}{
\begin{tabular}{lccccccccccccc} \toprule
 & {Venue} & \rotatebox{90}{ImageNet} & \rotatebox{90}{EuroSAT} & \rotatebox{90}{Caltech101} & \rotatebox{90}{OxfordPets} & \rotatebox{90}{UCF101} & \rotatebox{90}{DTD} & \rotatebox{90}{Flowers102} & \rotatebox{90}{SUN397} & \rotatebox{90}{RESISC45} & \rotatebox{90}{CIFAR10} & \rotatebox{90}{CIFAR100} & \rotatebox{90}{Average} \\ 
 \toprule
\multicolumn{14}{l}{Few-Shot Methods} \\ \midrule
CoOp (1-Shot) \cite{zhou2022learning} & IJCV (`22) & 60.6 & 58.4 & 91.7 & - & 63.8 & 40.1 & 71.2 & 64.1 & - & 83 & 55.6 & - \\
CoOp (5-Shot) \cite{zhou2022learning} & IJCV (`22) & 61.3 & 71.8 & 93.2 & - & 74.3 & 41.1 & 85.8 & 67.3 & - & 86.6 & 63.2 & - \\
CoOp (10-Shot) \cite{zhou2022learning} & IJCV (`22) & 62.3 & 81.6 & 94.6 & - & 77.2 & 65.8 & 92.1 & 69 & - & 88.5 & 66.6 & - \\
\toprule
\multicolumn{14}{l}{Label-Free Methods} \\ \hline
CLIP \cite{radford2021learning} & ICML (`21) & 61.9 & 40.6 & 90.5 & 85.0 & 61.0 & 42.9 & 66.6 & 60.8 & 49.8 & 88.8 & 64.2 & 64.7 \\
CuPL \cite{pratt2023does} & ICCV (`23) & 63.4 & 62.2 & 90.6 & 87.2 & 63.9 & 48.0 & 71.5 & 66.0 & 61.9 & 89.2 & 65.8 & 70.0 \\
MetaPrompt \cite{mirza2024meta} & ECCV (`24) & 65.0 & 55.6 & 92.9 & 88.1 & 67.9 & 50.8 & 73.9 & 67.0 & 64.0 & 89.9 & 66.3 & 71.0 \\
LaFTer \cite{mirza2024lafter} & NeurIPS (`24) & 64.2 & \textbf{73.9} & 93.3 & 82.7 & 68.2 & 46.1 & 71.0 & 64.5 & 68.3 & \textbf{95.8} & 74.6 & 73.0 \\
ProText \cite{khattak2024learning} & Arxiv & 64.9 & 51.4 & 93.4 & 89.0 & 66.4 & 50.7 & 74.2 & 66.8 & 57.4 & 89.5 & 66.1 & 70.0 \\
WaffleCLIP \cite{roth2023waffling} & ICCV (`23) & 63.5 & 46.7 & \textbf{94.8} & 88.1 & 65.8 & 51.0 & 68.7 & 65.6 & 63.4 & 90.9 & 67.2 & 69.6 \\
\rowcolor[HTML]{D9D9D9} 
Ours & & \textbf{65.4} & 73.5 & \textbf{94.8} & \textbf{89.3} & \textbf{68.3} & \textbf{56.1} & \textbf{82.7} & \textbf{67.0} & \textbf{75.4} & 94.9 & \textbf{75.6} & \textbf{76.6} \\ \bottomrule
\end{tabular}}
\caption{Top-1 accuracy (\%) for 11 datasets by using ViT-B/32
CLIP variant}
\label{tab:vitb32_main_results}
\end{table*}

%% file: tables/main_table_16.tex
\begin{table}[h!]
\centering
\renewcommand{\arraystretch}{1.3} 
\resizebox{\columnwidth}{!}{
\begin{tabular}{lccccccc}
\hline
 & \rotatebox{90}{EuroSAT} & \rotatebox{90}{Caltech101} & \rotatebox{90}{OxfordPets} & \rotatebox{90}{Flowers102} & \rotatebox{90}{SUN397} & \rotatebox{90}{CIFAR100} & \rotatebox{90}{Average} \\ \hline
AdaptCLIPZS (16-shot) \cite{saha2024improved} & 81.8 & 95.3 & 93.7 & 81.3 & 72.1 & 73.9 & 83.0 \\ \hline
LaFTer \cite{mirza2024lafter} & 72.1 & 94.3 & 81.5 & 65.9 & 65.9 & 76.3 & 76.0 \\
\rowcolor[HTML]{D9D9D9} 
Ours & \textbf{79.1} & \textbf{95.3} & \textbf{91.7} & \textbf{84.3} & \textbf{69.3} & \textbf{77.5} & \textbf{82.9} \\ \hline
\end{tabular}}
\caption{We compare our methodology with CLIP ViT-B/16 variant across six datasets on few-shot (AdaptCLIPZS) and label-free (LaFTer) methods. }
\label{tab:vitb16_main_results}
\end{table}

%% file: tables/ablation_table_ind_comp.tex
\begin{table}[H]
    \caption{\textnormal{\textbf{Ablation of integration of individual components that makes up the NoLA framework.} The top-1 accuracy for each variant is averaged across six datasets for their respective test sets.}}
    \centering
    \begin{tabular}{l c}
    \toprule
     & Avg. Top-1 Acc. \\
    \midrule
    CLIP zero-shot & 67.9 \\
    (+) CDE classifier & 72.0 \\
    (+) DL network & 73.9 \\ 
    (+) DINO-assisted Prompt Learning & {80.5} \\
    \bottomrule
    \end{tabular}
    \label{tab:ablation1}
\end{table}

%% file: tables/ablation_combine.tex
\begin{table}[H]
    \caption{\textnormal{\textbf{Alternate design choices:} Ablation 1 refers to the ablation of using CLIP vision encoder in the DL network of (b) in Figure \ref{NOLA_arc}. Ablation 2 refers to the ablation of using the CDE classifier as the pseudo-labeller of (c) in Figure \ref{NOLA_arc}. 
    The top-1 accuracy for each ablation is averaged across six datasets for their respective test sets.}}
    \small
    \setlength{\tabcolsep}{1pt}
    \centering
    \begin{tabular}{l c}
    \toprule
      Method & Avg. Top-1 Acc. \\
    \midrule
   Ours: NoLA  (using DINO based DL) & 80.5 \\
   
     Ablation 1: Replace DINO with CLIP in DL  & {77.8} \\
     Ablation 2: Replace DL with CDE & {76.2} \\

    \bottomrule
    \end{tabular}
    \label{tab:ablation_combined}
\end{table}

%% file: tables/suppl_3.tex
\begin{table}[!htb]
    \centering 
    \caption{\textnormal{\textbf{Ablation of the trainable components in NoLA.} The Top-1 accuracy is averaged across six datasets, namingly, EuroSAT, Caltech101, Oxford-pets, Flowers-102, SUN397, CIFAR100.}}
    \resizebox{\columnwidth}{!}{ 
    \begin{tabular}{lccc}
    \toprule
     Trainable components $\rightarrow$ & Prompts & CDE & Avg. Top-1 Acc. \\
     \midrule
    Setting 1 & \xmark & \xmark & {77.9} \\
    Setting 2 & \checkmark & \xmark & {78.2} \\
    Setting 3 & \xmark & \checkmark & {79.6} \\
    \textbf{NoLA} & \checkmark & \checkmark & {\textbf{80.5}} \\
    \bottomrule
    \end{tabular}}
    \label{tab:suppl_3}
\end{table}

%% file: tables/ablation_table_llm.tex

\begin{table}[H]
\setlength{\tabcolsep}{4pt}
    \centering 
    \caption{\textnormal{Comparison of NoLA's performance with GPT-3.5 descriptions and GPT-4o descriptions over six datasets.}}
    \small 
    \begin{tabular}{lccccccc}
    \toprule
    LLM & \rotatebox{90}{EuroSAT} & \rotatebox{90}{Caltech101} & \rotatebox{90}{Oxford-pets} & \rotatebox{90}{Flowers-102} & \rotatebox{90}{SUN397} & \rotatebox{90}{CIFAR100} & \rotatebox{90}{Average} \\
    \midrule
    GPT-3.5 & 73.47 & 94.84 & \textbf{89.34} & 82.70 & \textbf{67.02} & \textbf{75.62 }& 80.50 \\
    GPT-4o & \textbf{74.75} &\textbf{95.01} & 88.66 & \textbf{85.91} & 66.69 & 75.20 & \textbf{81.04} \\
    \bottomrule
    \end{tabular}
    \label{tab:gpt4omini}
\end{table}

%% file: tables/suppl_1.tex
\begin{table}[!htb]

    \centering 
    \caption{\textnormal{\textbf{The list of hyperparameters used to optimize alignment module $\bm{h}$ within the DL network.}}}
    \small 
    \begin{tabular}{lc}
    \toprule
    Hyperparameters & Value \\
    \midrule
    GPU & Nvidia A100 80GB \\
    Backbone & Pretrained DINO ViT B/16 \\
    Pretrained & ImageNet \\
    Input Size & 224x224 \\
    Epochs & 50 \\
    Optimizer & AdamW \\
    Learning Rate & $1e^{-3}$ \\
    Batch Size & 32 \\
    Samples trained on & Top-$k$ strategy \\
    Loss & Smoothed Cross-Entropy \\
    \bottomrule
    \end{tabular}
    \label{tab:suppl_1}
\end{table}

%% file: tables/suppl_2.tex
\begin{table}[H]
\setlength{\tabcolsep}{2pt}

    \centering 
    \caption{\textnormal{\textbf{The list of hyperparameters used for DINO-assisted prompt learning stage in the NoLA framework.}}}
    \small 
    \begin{tabular}{lc}
    \toprule
    Hyperparameters & Value \\
    \midrule
    GPU & Nvidia A100 80GB \\
    Backbone & ViT B/32 \\
    Input Size & 224x224 \\
    Prompt Tuning Method & VPT\cite{jia2022visual} \\
    Learnable Tokens & 16 \\
    Batch Size & 512 \\
    Optimizer & Adam \\
    Learning Rate & $4e^{-3}$ \\
    Loss & Smoothed Cross Entropy \\

    \bottomrule
    \end{tabular}
    \label{tab:suppl_2}
\end{table}

%% file: main.bbl
\begin{thebibliography}{81}
\providecommand{\natexlab}[1]{#1}

\bibitem[{Akiva, Purri, and Leotta(2022)}]{akiva2022self}
Akiva, P.; Purri, M.; and Leotta, M. 2022.
\newblock Self-supervised material and texture representation learning for remote sensing tasks.
\newblock In \emph{Proceedings of the IEEE/CVF Conference on Computer Vision and Pattern Recognition}, 8203--8215.

\bibitem[{Alom et~al.(2018)Alom, Taha, Yakopcic, Westberg, Sidike, Nasrin, Van~Esesn, Awwal, and Asari}]{alom2018history}
Alom, M.~Z.; Taha, T.~M.; Yakopcic, C.; Westberg, S.; Sidike, P.; Nasrin, M.~S.; Van~Esesn, B.~C.; Awwal, A. A.~S.; and Asari, V.~K. 2018.
\newblock The history began from alexnet: A comprehensive survey on deep learning approaches.
\newblock \emph{arXiv preprint arXiv:1803.01164}.

\bibitem[{Assran et~al.(2021)Assran, Caron, Misra, Bojanowski, Joulin, Ballas, and Rabbat}]{assran2021semi}
Assran, M.; Caron, M.; Misra, I.; Bojanowski, P.; Joulin, A.; Ballas, N.; and Rabbat, M. 2021.
\newblock Semi-supervised learning of visual features by non-parametrically predicting view assignments with support samples.
\newblock In \emph{Proceedings of the IEEE/CVF International Conference on Computer Vision}, 8443--8452.

\bibitem[{Bazi et~al.(2022)Bazi, Al~Rahhal, Mekhalfi, Al~Zuair, and Melgani}]{bazi2022bi}
Bazi, Y.; Al~Rahhal, M.~M.; Mekhalfi, M.~L.; Al~Zuair, M.~A.; and Melgani, F. 2022.
\newblock Bi-modal transformer-based approach for visual question answering in remote sensing imagery.
\newblock \emph{IEEE Transactions on Geoscience and Remote Sensing}, 60: 1--11.

\bibitem[{Berthelot et~al.(2019)Berthelot, Carlini, Goodfellow, Papernot, Oliver, and Raffel}]{berthelot2019mixmatch}
Berthelot, D.; Carlini, N.; Goodfellow, I.; Papernot, N.; Oliver, A.; and Raffel, C.~A. 2019.
\newblock Mixmatch: A holistic approach to semi-supervised learning.
\newblock \emph{Advances in neural information processing systems}, 32.

\bibitem[{Caron et~al.(2021)Caron, Touvron, Misra, J{\'e}gou, Mairal, Bojanowski, and Joulin}]{caron2021emerging}
Caron, M.; Touvron, H.; Misra, I.; J{\'e}gou, H.; Mairal, J.; Bojanowski, P.; and Joulin, A. 2021.
\newblock Emerging properties in self-supervised vision transformers.
\newblock In \emph{Proceedings of the IEEE/CVF international conference on computer vision}, 9650--9660.

\bibitem[{Chen et~al.(2020)Chen, Kornblith, Norouzi, and Hinton}]{chen2020simpleframeworkcontrastivelearning}
Chen, T.; Kornblith, S.; Norouzi, M.; and Hinton, G. 2020.
\newblock A Simple Framework for Contrastive Learning of Visual Representations.
\newblock arXiv:2002.05709.

\bibitem[{Chen and He(2021)}]{chen2021exploring}
Chen, X.; and He, K. 2021.
\newblock Exploring simple siamese representation learning.
\newblock In \emph{Proceedings of the IEEE/CVF conference on computer vision and pattern recognition}, 15750--15758.

\bibitem[{Chen et~al.(2023)Chen, Huang, Li, Xiong, and Lu}]{chen2023multiscale}
Chen, Y.; Huang, J.; Li, X.; Xiong, S.; and Lu, X. 2023.
\newblock Multiscale Salient Alignment Learning for Remote Sensing Image-Text Retrieval.
\newblock \emph{IEEE Transactions on Geoscience and Remote Sensing}.

\bibitem[{Cheng, Han, and Lu(2017)}]{cheng2017remote}
Cheng, G.; Han, J.; and Lu, X. 2017.
\newblock Remote sensing image scene classification: Benchmark and state of the art.
\newblock \emph{Proceedings of the IEEE}, 105(10): 1865--1883.

\bibitem[{Cimpoi et~al.(2014)Cimpoi, Maji, Kokkinos, Mohamed, and Vedaldi}]{cimpoi2014describing}
Cimpoi, M.; Maji, S.; Kokkinos, I.; Mohamed, S.; and Vedaldi, A. 2014.
\newblock Describing textures in the wild.
\newblock In \emph{Proceedings of the IEEE conference on computer vision and pattern recognition}, 3606--3613.

\bibitem[{Cozzolino et~al.(2024)Cozzolino, Poggi, Corvi, Nie{\ss}ner, and Verdoliva}]{cozzolino2024raising}
Cozzolino, D.; Poggi, G.; Corvi, R.; Nie{\ss}ner, M.; and Verdoliva, L. 2024.
\newblock Raising the Bar of AI-generated Image Detection with CLIP.
\newblock In \emph{Proceedings of the IEEE/CVF Conference on Computer Vision and Pattern Recognition}, 4356--4366.

\bibitem[{Deng et~al.(2009)Deng, Dong, Socher, Li, Li, and Fei-Fei}]{deng2009imagenet}
Deng, J.; Dong, W.; Socher, R.; Li, L.-J.; Li, K.; and Fei-Fei, L. 2009.
\newblock Imagenet: A large-scale hierarchical image database.
\newblock In \emph{2009 IEEE conference on computer vision and pattern recognition}, 248--255. Ieee.

\bibitem[{Dosovitskiy et~al.(2020)Dosovitskiy, Beyer, Kolesnikov, Weissenborn, Zhai, Unterthiner, Dehghani, Minderer, Heigold, Gelly et~al.}]{dosovitskiy2020image}
Dosovitskiy, A.; Beyer, L.; Kolesnikov, A.; Weissenborn, D.; Zhai, X.; Unterthiner, T.; Dehghani, M.; Minderer, M.; Heigold, G.; Gelly, S.; et~al. 2020.
\newblock An image is worth 16x16 words: Transformers for image recognition at scale.
\newblock \emph{arXiv preprint arXiv:2010.11929}.

\bibitem[{Eldele et~al.(2023)Eldele, Ragab, Chen, Wu, Kwoh, Li, and Guan}]{eldele2023self}
Eldele, E.; Ragab, M.; Chen, Z.; Wu, M.; Kwoh, C.-K.; Li, X.; and Guan, C. 2023.
\newblock Self-supervised contrastive representation learning for semi-supervised time-series classification.
\newblock \emph{IEEE Transactions on Pattern Analysis and Machine Intelligence}.

\bibitem[{Fei-Fei, Fergus, and Perona(2006)}]{fei2006one}
Fei-Fei, L.; Fergus, R.; and Perona, P. 2006.
\newblock One-shot learning of object categories.
\newblock \emph{IEEE transactions on pattern analysis and machine intelligence}, 28(4): 594--611.

\bibitem[{Gao et~al.(2024)Gao, Geng, Zhang, Ma, Fang, Zhang, Li, and Qiao}]{gao2024clip}
Gao, P.; Geng, S.; Zhang, R.; Ma, T.; Fang, R.; Zhang, Y.; Li, H.; and Qiao, Y. 2024.
\newblock Clip-adapter: Better vision-language models with feature adapters.
\newblock \emph{International Journal of Computer Vision}, 132(2): 581--595.

\bibitem[{Helber et~al.(2019)Helber, Bischke, Dengel, and Borth}]{helber2019eurosat}
Helber, P.; Bischke, B.; Dengel, A.; and Borth, D. 2019.
\newblock Eurosat: A novel dataset and deep learning benchmark for land use and land cover classification.
\newblock \emph{IEEE Journal of Selected Topics in Applied Earth Observations and Remote Sensing}, 12(7): 2217--2226.

\bibitem[{Hou et~al.(2024)Hou, Chen, Chen, Hong, Wang, Feng, Khan, Khan, and You}]{hou2024visual}
Hou, W.; Chen, S.; Chen, S.; Hong, Z.; Wang, Y.; Feng, X.; Khan, S.; Khan, F.~S.; and You, X. 2024.
\newblock Visual-Augmented Dynamic Semantic Prototype for Generative Zero-Shot Learning.
\newblock In \emph{Proceedings of the IEEE/CVF Conference on Computer Vision and Pattern Recognition}, 23627--23637.

\bibitem[{Hoyer et~al.(2023)Hoyer, Tan, Naeem, Van~Gool, and Tombari}]{hoyer2023semivl}
Hoyer, L.; Tan, D.~J.; Naeem, M.~F.; Van~Gool, L.; and Tombari, F. 2023.
\newblock SemiVL: Semi-Supervised Semantic Segmentation with Vision-Language Guidance.
\newblock \emph{arXiv preprint arXiv:2311.16241}.

\bibitem[{Huang, Chu, and Wei(2022)}]{huang2022unsupervised}
Huang, T.; Chu, J.; and Wei, F. 2022.
\newblock Unsupervised prompt learning for vision-language models.
\newblock \emph{arXiv preprint arXiv:2204.03649}.

\bibitem[{Jia et~al.(2021)Jia, Yang, Xia, Chen, Parekh, Pham, Le, Sung, Li, and Duerig}]{jia2021scaling}
Jia, C.; Yang, Y.; Xia, Y.; Chen, Y.-T.; Parekh, Z.; Pham, H.; Le, Q.; Sung, Y.-H.; Li, Z.; and Duerig, T. 2021.
\newblock Scaling up visual and vision-language representation learning with noisy text supervision.
\newblock In \emph{International conference on machine learning}, 4904--4916. PMLR.

\bibitem[{Jia et~al.(2022)Jia, Tang, Chen, Cardie, Belongie, Hariharan, and Lim}]{jia2022visual}
Jia, M.; Tang, L.; Chen, B.-C.; Cardie, C.; Belongie, S.; Hariharan, B.; and Lim, S.-N. 2022.
\newblock Visual prompt tuning.
\newblock In \emph{ECCV}, 709--727. Springer.

\bibitem[{Joo et~al.(2023)Joo, Vo, Yamazaki, and Le}]{joo2023clip}
Joo, H.~K.; Vo, K.; Yamazaki, K.; and Le, N. 2023.
\newblock Clip-tsa: Clip-assisted temporal self-attention for weakly-supervised video anomaly detection.
\newblock In \emph{2023 IEEE International Conference on Image Processing (ICIP)}, 3230--3234. IEEE.

\bibitem[{Khattak et~al.(2024)Khattak, Naeem, Naseer, Van~Gool, and Tombari}]{khattak2024learning}
Khattak, M.~U.; Naeem, M.~F.; Naseer, M.; Van~Gool, L.; and Tombari, F. 2024.
\newblock Learning to Prompt with Text Only Supervision for Vision-Language Models.
\newblock \emph{arXiv preprint arXiv:2401.02418}.

\bibitem[{Khattak et~al.(2023)Khattak, Rasheed, Maaz, Khan, and Khan}]{khattak2023maple}
Khattak, M.~U.; Rasheed, H.; Maaz, M.; Khan, S.; and Khan, F.~S. 2023.
\newblock Maple: Multi-modal prompt learning.
\newblock In \emph{Proceedings of the IEEE/CVF Conference on Computer Vision and Pattern Recognition}, 19113--19122.

\bibitem[{Kingma and Ba(2014)}]{kingma2014adam}
Kingma, D.~P.; and Ba, J. 2014.
\newblock Adam: A method for stochastic optimization.
\newblock \emph{arXiv preprint arXiv:1412.6980}.

\bibitem[{Ko{\c{c}}yi{\u{g}}it, Hospedales, and Bilen(2023)}]{koccyiugit2023accelerating}
Ko{\c{c}}yi{\u{g}}it, M.~T.; Hospedales, T.~M.; and Bilen, H. 2023.
\newblock Accelerating Self-Supervised Learning via Efficient Training Strategies.
\newblock In \emph{Proceedings of the IEEE/CVF Winter Conference on Applications of Computer Vision}, 5654--5664.

\bibitem[{Krizhevsky, Hinton et~al.(2009)}]{krizhevsky2009learning}
Krizhevsky, A.; Hinton, G.; et~al. 2009.
\newblock Learning multiple layers of features from tiny images.

\bibitem[{Krizhevsky, Sutskever, and Hinton(2012)}]{krizhevsky2012imagenet}
Krizhevsky, A.; Sutskever, I.; and Hinton, G.~E. 2012.
\newblock Imagenet classification with deep convolutional neural networks.
\newblock \emph{Advances in neural information processing systems}, 25.

\bibitem[{Kurakin et~al.(2020)Kurakin, Raffel, Berthelot, Cubuk, Zhang, Sohn, and Carlini}]{kurakin2020remixmatch}
Kurakin, A.; Raffel, C.; Berthelot, D.; Cubuk, E.~D.; Zhang, H.; Sohn, K.; and Carlini, N. 2020.
\newblock Remixmatch: Semi-supervised learning with distribution matching and augmentation anchoring.

\bibitem[{Lee et~al.(2023)Lee, Tsai, Chiu, and Lee}]{lee2023multimodal}
Lee, Y.-L.; Tsai, Y.-H.; Chiu, W.-C.; and Lee, C.-Y. 2023.
\newblock Multimodal Prompting with Missing Modalities for Visual Recognition.
\newblock In \emph{Proceedings of the IEEE/CVF Conference on Computer Vision and Pattern Recognition}, 14943--14952.

\bibitem[{Lester, Al-Rfou, and Constant(2021)}]{lester2021power}
Lester, B.; Al-Rfou, R.; and Constant, N. 2021.
\newblock The power of scale for parameter-efficient prompt tuning.
\newblock \emph{arXiv preprint arXiv:2104.08691}.

\bibitem[{Li et~al.(2022)Li, Li, Xiong, and Hoi}]{li2022blip}
Li, J.; Li, D.; Xiong, C.; and Hoi, S. 2022.
\newblock Blip: Bootstrapping language-image pre-training for unified vision-language understanding and generation.
\newblock In \emph{International conference on machine learning}, 12888--12900. PMLR.

\bibitem[{Li, Li, and Wang(2023)}]{li2023class}
Li, M.; Li, Q.; and Wang, Y. 2023.
\newblock Class Balanced Adaptive Pseudo Labeling for Federated Semi-Supervised Learning.
\newblock In \emph{Proceedings of the IEEE/CVF Conference on Computer Vision and Pattern Recognition}, 16292--16301.

\bibitem[{Li et~al.(2023{\natexlab{a}})Li, Wen, Hu, and Zhou}]{li2023rs}
Li, X.; Wen, C.; Hu, Y.; and Zhou, N. 2023{\natexlab{a}}.
\newblock Rs-clip: Zero shot remote sensing scene classification via contrastive vision-language supervision.
\newblock \emph{International Journal of Applied Earth Observation and Geoinformation}, 124: 103497.

\bibitem[{Li et~al.(2023{\natexlab{b}})Li, Qi, Shi, and Gao}]{li2023iomatch}
Li, Z.; Qi, L.; Shi, Y.; and Gao, Y. 2023{\natexlab{b}}.
\newblock IOMatch: Simplifying open-set semi-supervised learning with joint inliers and outliers utilization.
\newblock In \emph{Proceedings of the IEEE/CVF International Conference on Computer Vision}, 15870--15879.

\bibitem[{Liang et~al.(2023)Liang, Wu, Dai, Li, Zhao, Zhang, Zhang, Vajda, and Marculescu}]{liang2023open}
Liang, F.; Wu, B.; Dai, X.; Li, K.; Zhao, Y.; Zhang, H.; Zhang, P.; Vajda, P.; and Marculescu, D. 2023.
\newblock Open-vocabulary semantic segmentation with mask-adapted clip.
\newblock In \emph{Proceedings of the IEEE/CVF Conference on Computer Vision and Pattern Recognition}, 7061--7070.

\bibitem[{Liu et~al.(2023)Liu, Zhang, Lin, Zhang, Tan, Han, Li, Ding, and Wang}]{liu2023ambiguity}
Liu, C.; Zhang, W.; Lin, X.; Zhang, W.; Tan, X.; Han, J.; Li, X.; Ding, E.; and Wang, J. 2023.
\newblock Ambiguity-Resistant Semi-Supervised Learning for Dense Object Detection.
\newblock In \emph{Proceedings of the IEEE/CVF Conference on Computer Vision and Pattern Recognition}, 15579--15588.

\bibitem[{Liu et~al.(2024)Liu, Li, Wu, and Lee}]{liu2024visual}
Liu, H.; Li, C.; Wu, Q.; and Lee, Y.~J. 2024.
\newblock Visual instruction tuning.
\newblock \emph{Advances in neural information processing systems}, 36.

\bibitem[{Lu et~al.(2022)Lu, Liu, Zhang, Liu, and Tian}]{lu2022prompt}
Lu, Y.; Liu, J.; Zhang, Y.; Liu, Y.; and Tian, X. 2022.
\newblock Prompt distribution learning.
\newblock In \emph{Proceedings of the IEEE/CVF Conference on Computer Vision and Pattern Recognition}, 5206--5215.

\bibitem[{Mirza et~al.(2024{\natexlab{a}})Mirza, Karlinsky, Lin, Doveh, , Micorek, Kozinski, Kuhene, and Possegger}]{mirza2024meta}
Mirza, M.~J.; Karlinsky, L.; Lin, W.; Doveh, S.; ; Micorek, J.; Kozinski, M.; Kuhene, H.; and Possegger, H. 2024{\natexlab{a}}.
\newblock {Meta-Prompting for Automating Zero-shot Visual Recognition with LLMs}.
\newblock In \emph{Proceedings of the European Conference on Computer Vision (ECCV)}.

\bibitem[{Mirza et~al.(2024{\natexlab{b}})Mirza, Karlinsky, Lin, Possegger, Kozinski, Feris, and Bischof}]{mirza2024lafter}
Mirza, M.~J.; Karlinsky, L.; Lin, W.; Possegger, H.; Kozinski, M.; Feris, R.; and Bischof, H. 2024{\natexlab{b}}.
\newblock Lafter: Label-free tuning of zero-shot classifier using language and unlabeled image collections.
\newblock \emph{Advances in Neural Information Processing Systems}, 36.

\bibitem[{Naeem et~al.(2023{\natexlab{a}})Naeem, Khan, Xian, Afzal, Stricker, Van~Gool, and Tombari}]{naeem2023i2mvformer}
Naeem, M.~F.; Khan, M. G. Z.~A.; Xian, Y.; Afzal, M.~Z.; Stricker, D.; Van~Gool, L.; and Tombari, F. 2023{\natexlab{a}}.
\newblock I2mvformer: Large language model generated multi-view document supervision for zero-shot image classification.
\newblock In \emph{Proceedings of the IEEE/CVF Conference on Computer Vision and Pattern Recognition}, 15169--15179.

\bibitem[{Naeem et~al.(2023{\natexlab{b}})Naeem, Xian, Zhai, Hoyer, Van~Gool, and Tombari}]{naeem2023silc}
Naeem, M.~F.; Xian, Y.; Zhai, X.; Hoyer, L.; Van~Gool, L.; and Tombari, F. 2023{\natexlab{b}}.
\newblock Silc: Improving vision language pretraining with self-distillation.
\newblock \emph{arXiv preprint arXiv:2310.13355}.

\bibitem[{Nguyen and Yang(2023)}]{nguyen2023boosting}
Nguyen, K.-B.; and Yang, J.-S. 2023.
\newblock Boosting Semi-Supervised Learning by bridging high and low-confidence predictions.
\newblock In \emph{Proceedings of the IEEE/CVF International Conference on Computer Vision}, 1028--1038.

\bibitem[{Nilsback and Zisserman(2008)}]{nilsback2008automated}
Nilsback, M.-E.; and Zisserman, A. 2008.
\newblock Automated flower classification over a large number of classes.
\newblock In \emph{2008 Sixth Indian conference on computer vision, graphics \& image processing}, 722--729. IEEE.

\bibitem[{Pan et~al.(2024)Pan, Yaman, Velipasalar, and Ren}]{pan2024clip}
Pan, C.; Yaman, B.; Velipasalar, S.; and Ren, L. 2024.
\newblock Clip-bevformer: Enhancing multi-view image-based bev detector with ground truth flow.
\newblock In \emph{Proceedings of the IEEE/CVF Conference on Computer Vision and Pattern Recognition}, 15216--15225.

\bibitem[{Pantazis et~al.(2022)Pantazis, Brostow, Jones, and Mac~Aodha}]{pantazis2022svl}
Pantazis, O.; Brostow, G.; Jones, K.; and Mac~Aodha, O. 2022.
\newblock Svl-adapter: Self-supervised adapter for vision-language pretrained models.
\newblock \emph{arXiv preprint arXiv:2210.03794}.

\bibitem[{Park and Van~Hentenryck(2023)}]{park2023self}
Park, S.; and Van~Hentenryck, P. 2023.
\newblock Self-supervised primal-dual learning for constrained optimization.
\newblock In \emph{Proceedings of the AAAI Conference on Artificial Intelligence}, volume~37, 4052--4060.

\bibitem[{Parkhi et~al.(2012)Parkhi, Vedaldi, Zisserman, and Jawahar}]{parkhi2012cats}
Parkhi, O.~M.; Vedaldi, A.; Zisserman, A.; and Jawahar, C. 2012.
\newblock Cats and dogs.
\newblock In \emph{2012 IEEE conference on computer vision and pattern recognition}, 3498--3505. IEEE.

\bibitem[{Pourpanah et~al.(2022)Pourpanah, Abdar, Luo, Zhou, Wang, Lim, Wang, and Wu}]{pourpanah2022review}
Pourpanah, F.; Abdar, M.; Luo, Y.; Zhou, X.; Wang, R.; Lim, C.~P.; Wang, X.-Z.; and Wu, Q.~J. 2022.
\newblock A review of generalized zero-shot learning methods.
\newblock \emph{IEEE transactions on pattern analysis and machine intelligence}, 45(4): 4051--4070.

\bibitem[{Pratt et~al.(2023)Pratt, Covert, Liu, and Farhadi}]{pratt2023does}
Pratt, S.; Covert, I.; Liu, R.; and Farhadi, A. 2023.
\newblock What does a platypus look like? generating customized prompts for zero-shot image classification.
\newblock In \emph{Proceedings of the IEEE/CVF International Conference on Computer Vision}, 15691--15701.

\bibitem[{Qiu et~al.(2022)Qiu, Yu, Yi, Guan, Shi, and Tong}]{qiu2022open}
Qiu, C.; Yu, A.; Yi, X.; Guan, N.; Shi, D.; and Tong, X. 2022.
\newblock Open Self-Supervised Features for Remote-Sensing Image Scene Classification Using Very Few Samples.
\newblock \emph{IEEE Geoscience and Remote Sensing Letters}, 20: 1--5.

\bibitem[{Radford et~al.(2021)Radford, Kim, Hallacy, Ramesh, Goh, Agarwal, Sastry, Askell, Mishkin, Clark et~al.}]{radford2021learning}
Radford, A.; Kim, J.~W.; Hallacy, C.; Ramesh, A.; Goh, G.; Agarwal, S.; Sastry, G.; Askell, A.; Mishkin, P.; Clark, J.; et~al. 2021.
\newblock Learning transferable visual models from natural language supervision.
\newblock In \emph{International conference on machine learning}, 8748--8763. PMLR.

\bibitem[{Roth et~al.(2023)Roth, Kim, Koepke, Vinyals, Schmid, and Akata}]{roth2023waffling}
Roth, K.; Kim, J.~M.; Koepke, A.; Vinyals, O.; Schmid, C.; and Akata, Z. 2023.
\newblock Waffling around for performance: Visual classification with random words and broad concepts.
\newblock In \emph{Proceedings of the IEEE/CVF International Conference on Computer Vision}, 15746--15757.

\bibitem[{Saha, Van~Horn, and Maji(2024)}]{saha2024improved}
Saha, O.; Van~Horn, G.; and Maji, S. 2024.
\newblock Improved Zero-Shot Classification by Adapting VLMs with Text Descriptions.
\newblock In \emph{Proceedings of the IEEE/CVF Conference on Computer Vision and Pattern Recognition}, 17542--17552.

\bibitem[{Schiappa, Rawat, and Shah(2023)}]{schiappa2023self}
Schiappa, M.~C.; Rawat, Y.~S.; and Shah, M. 2023.
\newblock Self-supervised learning for videos: A survey.
\newblock \emph{ACM Computing Surveys}, 55(13s): 1--37.

\bibitem[{Shu et~al.(2022)Shu, Nie, Huang, Yu, Goldstein, Anandkumar, and Xiao}]{shu2022test}
Shu, M.; Nie, W.; Huang, D.-A.; Yu, Z.; Goldstein, T.; Anandkumar, A.; and Xiao, C. 2022.
\newblock Test-time prompt tuning for zero-shot generalization in vision-language models.
\newblock \emph{Advances in Neural Information Processing Systems}, 35: 14274--14289.

\bibitem[{Sohn et~al.(2020)Sohn, Berthelot, Carlini, Zhang, Zhang, Raffel, Cubuk, Kurakin, and Li}]{sohn2020fixmatch}
Sohn, K.; Berthelot, D.; Carlini, N.; Zhang, Z.; Zhang, H.; Raffel, C.~A.; Cubuk, E.~D.; Kurakin, A.; and Li, C.-L. 2020.
\newblock Fixmatch: Simplifying semi-supervised learning with consistency and confidence.
\newblock \emph{Advances in neural information processing systems}, 33: 596--608.

\bibitem[{Soomro, Zamir, and Shah(2012)}]{soomro2012ucf101}
Soomro, K.; Zamir, A.~R.; and Shah, M. 2012.
\newblock UCF101: A dataset of 101 human actions classes from videos in the wild.
\newblock \emph{arXiv preprint arXiv:1212.0402}.

\bibitem[{Stojnic and Risojevic(2021)}]{stojnic2021self}
Stojnic, V.; and Risojevic, V. 2021.
\newblock Self-supervised learning of remote sensing scene representations using contrastive multiview coding.
\newblock In \emph{Proceedings of the IEEE/CVF Conference on Computer Vision and Pattern Recognition}, 1182--1191.

\bibitem[{Szegedy et~al.(2016)Szegedy, Vanhoucke, Ioffe, Shlens, and Wojna}]{szegedy2016rethinking}
Szegedy, C.; Vanhoucke, V.; Ioffe, S.; Shlens, J.; and Wojna, Z. 2016.
\newblock Rethinking the inception architecture for computer vision.
\newblock In \emph{Proceedings of the IEEE conference on computer vision and pattern recognition}, 2818--2826.

\bibitem[{Vaswani et~al.(2017)Vaswani, Shazeer, Parmar, Uszkoreit, Jones, Gomez, Kaiser, and Polosukhin}]{vaswani2017attention}
Vaswani, A.; Shazeer, N.; Parmar, N.; Uszkoreit, J.; Jones, L.; Gomez, A.~N.; Kaiser, {\L}.; and Polosukhin, I. 2017.
\newblock Attention is all you need.
\newblock \emph{Advances in neural information processing systems}, 30.

\bibitem[{Wei and Gan(2023)}]{wei2023towards}
Wei, T.; and Gan, K. 2023.
\newblock Towards Realistic Long-Tailed Semi-Supervised Learning: Consistency Is All You Need.
\newblock In \emph{Proceedings of the IEEE/CVF Conference on Computer Vision and Pattern Recognition}, 3469--3478.

\bibitem[{Wu et~al.(2024)Wu, Zhou, Pang, Zhou, Yan, Wang, and Zhang}]{wu2024vadclip}
Wu, P.; Zhou, X.; Pang, G.; Zhou, L.; Yan, Q.; Wang, P.; and Zhang, Y. 2024.
\newblock Vadclip: Adapting vision-language models for weakly supervised video anomaly detection.
\newblock In \emph{Proceedings of the AAAI Conference on Artificial Intelligence}, volume~38, 6074--6082.

\bibitem[{Wysocza{\'n}ska et~al.(2024)Wysocza{\'n}ska, Ramamonjisoa, Trzci{\'n}ski, and Sim{\'e}oni}]{wysoczanska2024clip}
Wysocza{\'n}ska, M.; Ramamonjisoa, M.; Trzci{\'n}ski, T.; and Sim{\'e}oni, O. 2024.
\newblock Clip-diy: Clip dense inference yields open-vocabulary semantic segmentation for-free.
\newblock In \emph{Proceedings of the IEEE/CVF Winter Conference on Applications of Computer Vision}, 1403--1413.

\bibitem[{Xiao et~al.(2016)Xiao, Ehinger, Hays, Torralba, and Oliva}]{xiao2016sun}
Xiao, J.; Ehinger, K.~A.; Hays, J.; Torralba, A.; and Oliva, A. 2016.
\newblock Sun database: Exploring a large collection of scene categories.
\newblock \emph{International Journal of Computer Vision}, 119: 3--22.

\bibitem[{Xu et~al.(2020)Xu, Xian, Wang, Schiele, and Akata}]{xu2020attribute}
Xu, W.; Xian, Y.; Wang, J.; Schiele, B.; and Akata, Z. 2020.
\newblock Attribute prototype network for zero-shot learning.
\newblock \emph{Advances in Neural Information Processing Systems}, 33: 21969--21980.

\bibitem[{Yan et~al.(2024)Yan, Wu, Qin, Han, Cui, and Li}]{yan2024universal}
Yan, Z.; Wu, Y.; Qin, Y.; Han, X.; Cui, S.; and Li, G. 2024.
\newblock Universal semi-supervised model adaptation via collaborative consistency training.
\newblock In \emph{Proceedings of the IEEE/CVF Winter Conference on Applications of Computer Vision}, 872--882.

\bibitem[{Yao et~al.(2021)Yao, Huang, Hou, Lu, Niu, Xu, Liang, Li, Jiang, and Xu}]{yao2021filip}
Yao, L.; Huang, R.; Hou, L.; Lu, G.; Niu, M.; Xu, H.; Liang, X.; Li, Z.; Jiang, X.; and Xu, C. 2021.
\newblock Filip: Fine-grained interactive language-image pre-training.
\newblock \emph{arXiv preprint arXiv:2111.07783}.

\bibitem[{You et~al.(2023)You, Gu, Ham, Park, Kim, Hong, Baek, and Roh}]{you2023cxr}
You, K.; Gu, J.; Ham, J.; Park, B.; Kim, J.; Hong, E.~K.; Baek, W.; and Roh, B. 2023.
\newblock Cxr-clip: Toward large scale chest x-ray language-image pre-training.
\newblock In \emph{International Conference on Medical Image Computing and Computer-Assisted Intervention}, 101--111. Springer.

\bibitem[{Yu et~al.(2022)Yu, Wang, Vasudevan, Yeung, Seyedhosseini, and Wu}]{yu2022coca}
Yu, J.; Wang, Z.; Vasudevan, V.; Yeung, L.; Seyedhosseini, M.; and Wu, Y. 2022.
\newblock Coca: Contrastive captioners are image-text foundation models.
\newblock \emph{arXiv preprint arXiv:2205.01917}.

\bibitem[{Yuan et~al.(2021{\natexlab{a}})Yuan, Chen, Chen, Codella, Dai, Gao, Hu, Huang, Li, Li et~al.}]{yuan2021florence}
Yuan, L.; Chen, D.; Chen, Y.-L.; Codella, N.; Dai, X.; Gao, J.; Hu, H.; Huang, X.; Li, B.; Li, C.; et~al. 2021{\natexlab{a}}.
\newblock Florence: A new foundation model for computer vision.
\newblock \emph{arXiv preprint arXiv:2111.11432}.

\bibitem[{Yuan, Zhan, and Xiong(2023)}]{yuan2023parameter}
Yuan, Y.; Zhan, Y.; and Xiong, Z. 2023.
\newblock Parameter-Efficient Transfer Learning for Remote Sensing Image-Text Retrieval.
\newblock \emph{IEEE Transactions on Geoscience and Remote Sensing}.

\bibitem[{Yuan et~al.(2021{\natexlab{b}})Yuan, Zhang, Rong, Li, Chen, Wang, Fu, and Sun}]{yuan2021lightweight}
Yuan, Z.; Zhang, W.; Rong, X.; Li, X.; Chen, J.; Wang, H.; Fu, K.; and Sun, X. 2021{\natexlab{b}}.
\newblock A lightweight multi-scale crossmodal text-image retrieval method in remote sensing.
\newblock \emph{IEEE Transactions on Geoscience and Remote Sensing}, 60: 1--19.

\bibitem[{Zhang et~al.(2024)Zhang, Xu, Qiu, Yan, Lang, and Zhou}]{zhang2024mediclip}
Zhang, X.; Xu, M.; Qiu, D.; Yan, R.; Lang, N.; and Zhou, X. 2024.
\newblock MediCLIP: Adapting CLIP for Few-shot Medical Image Anomaly Detection.
\newblock \emph{arXiv preprint arXiv:2405.11315}.

\bibitem[{Zhao et~al.(2023)Zhao, Liu, Wu, Li, Wang, Teng, Liu, Li, Cui, Wang et~al.}]{zhao2023clip}
Zhao, Z.; Liu, Y.; Wu, H.; Li, Y.; Wang, S.; Teng, L.; Liu, D.; Li, X.; Cui, Z.; Wang, Q.; et~al. 2023.
\newblock Clip in medical imaging: A comprehensive survey.
\newblock \emph{arXiv preprint arXiv:2312.07353}.

\bibitem[{Zhou et~al.(2022{\natexlab{a}})Zhou, Yang, Loy, and Liu}]{zhou2022conditional}
Zhou, K.; Yang, J.; Loy, C.~C.; and Liu, Z. 2022{\natexlab{a}}.
\newblock Conditional prompt learning for vision-language models.
\newblock In \emph{Proceedings of the IEEE/CVF Conference on Computer Vision and Pattern Recognition}, 16816--16825.

\bibitem[{Zhou et~al.(2022{\natexlab{b}})Zhou, Yang, Loy, and Liu}]{zhou2022learning}
Zhou, K.; Yang, J.; Loy, C.~C.; and Liu, Z. 2022{\natexlab{b}}.
\newblock Learning to prompt for vision-language models.
\newblock \emph{International Journal of Computer Vision}, 130(9): 2337--2348.

\bibitem[{Zhu, Liu, and Huang(2023)}]{zhu2023hnssl}
Zhu, W.; Liu, J.; and Huang, Y. 2023.
\newblock Hnssl: Hard negative-based self-supervised learning.
\newblock In \emph{Proceedings of the IEEE/CVF Conference on Computer Vision and Pattern Recognition}, 4777--4786.

\end{thebibliography}
